\documentclass{article}

\usepackage[preprint]{neurips_2020}
\usepackage[utf8]{inputenc} % allow utf-8 input
\usepackage[T1]{fontenc}    % use 8-bit T1 fonts
\usepackage[breaklinks]{hyperref}
\usepackage{url}            % simple URL typesetting
\usepackage{booktabs}       % professional-quality tables
\usepackage{amsfonts}       % blackboard math symbols
\usepackage{nicefrac}       % compact symbols for 1/2, etc.
\usepackage{microtype}      % microtypography
\usepackage{microtype}
\usepackage{graphicx}
\usepackage{booktabs} % for professional tables
\usepackage{amsmath}
\usepackage{cleveref}
\usepackage{algorithm}      % algorithm
\usepackage[noend]{algpseudocode} % algorithm
\newtheorem{assumption}{Assumption}
\newtheorem{theorem}{Theorem}
\newtheorem{corollary}{Corollary}
\newtheorem{definition}{Definition}

\usepackage{pgfplots}
\pgfplotsset{width=8cm,height=5cm,compat=1.9}
\newlength\figureheight
\newlength\figurewidth
\usepackage{subfig}
\usepackage{amsmath}
\usepackage{amssymb}
\usepackage{amsfonts}
\usepackage{amstext}
\usepackage{xcolor}
\usepackage{xspace}
\usepackage{appendix}

\makeatletter
\newcommand*{\addFileDependency}[1]{% argument=file name and extension
  \typeout{(#1)}
  \@addtofilelist{#1}
  \IfFileExists{#1}{}{\typeout{No file #1.}}
}
\makeatother

\newcommand*{\myexternaldocument}[1]{%
    \externaldocument{#1}%
    \addFileDependency{#1.tex}%
    \addFileDependency{#1.aux}%
}
\usepackage{xr}
\myexternaldocument{appendix}

\crefname{appsec}{Appendix}{Appendices}

\definecolor{mygreen}{rgb}{0.2, 0.7, 0.2}
\definecolor{myorange}{rgb}{0.9, 0.5, 0.0}

\definecolor{blue}{HTML}{5DA5DA}
\definecolor{orange}{HTML}{FAA43A} 
\definecolor{green}{HTML}{60BD68} 
\definecolor{pink}{HTML}{F17CB0} 
\definecolor{brown}{HTML}{B2912F} 
\definecolor{purple}{HTML}{B276B2} 
\definecolor{yellow}{HTML}{DECF3F} 
\definecolor{red}{HTML}{F15854} 
\definecolor{gray}{HTML}{4D4D4D} 

\newcommand{\name}[1]{{\textsc{#1}}\xspace}

\newcommand{\mcmc}{\name{mcmc}}

\newcommand{\sde}{\name{sde}}
\newcommand{\fpe}{\name{fpe}}
\newcommand{\sg}{\name{sg}}
\newcommand{\sgd}{\name{sgd}}
\newcommand{\sgmcmc}{\name{sg-mcmc}}
\newcommand{\sgld}{\name{sgld}}
\newcommand{\sgfs}{\name{sgfs}}
\newcommand{\sgrld}{\name{sgrld}}
\newcommand{\sghmc}{\name{sghmc}}
\newcommand{\sgrhmc}{\name{sgrhmc}}
\newcommand{\isgd}{\name{i-sgd}}
\newcommand{\isgda}{\name{i-sgd-x-a}}
\newcommand{\isgdb}{\name{i-sgd-x-b}}
\newcommand{\isgdc}{\name{i-sgd-x-c}}

\newcommand{\isgdGb}{\name{i-sgd-G-b}}
\newcommand{\isgdGc}{\name{i-sgd-G-c}}

\newcommand{\isgdalphaa}{\name{i-sgd-$\alpha$-a}}

\newcommand{\isgdalphac}{\name{i-sgd-$\alpha$-c}}

\newcommand{\isgdAlpha}{\name{\textbf{i-sgd-}$\boldsymbol{\alpha}$}}

\newcommand{\isgdAlphac}{\name{\textbf{i-sgd-}$\boldsymbol{\alpha}$-c}}

\newcommand{\isgdalpha}{\name{i-sgd-$\alpha$}}
\newcommand{\isgdG}{\name{i-sgd-G}}

\newcommand{\isgdx}{\name{i-sgd-x}}

\newcommand{\isgdALL}{\name{i-sgd-$\alpha$/G-a/b/c}}

\newcommand{\slr}{\name{(slr)}}

\newcommand{\mcd}{\name{mcd}}
\newcommand{\swag}{\name{swag}}

\newcommand{\mandt}{\name{v-sgd}}
\newcommand{\cyc}{\name{csghmc}}

\newcommand{\mnll}{\name{mnll}}
\newcommand{\rmse}{\name{rmse}}
\newcommand{\acc}{\name{acc}}

\newcommand{\mnist}{\textsc{mnist }}
\newcommand{\cifar}{\textsc{cifar10 }}

\newcommand{\uci}{\textsc{uci }}

\newcommand{\wvect}{\mathbf{w}}

\newcommand{\zerovect}{\mathbf{0}}

\newcommand{\thetavect}{\boldsymbol{\theta}}

\newcommand{\zetavect}{\boldsymbol{z}}

\newcommand{\Lambdamath}{\boldsymbol{\Lambda}}
\newcommand{\Bmath}{\boldsymbol{B}(\boldsymbol{\theta})}
\newcommand{\Cmath}{\boldsymbol{C}(\boldsymbol{\theta})}
\newcommand{\Sigmamath}{\boldsymbol{\Sigma}(\boldsymbol{\theta})}
\newcommand{\Amath}{\boldsymbol{A}(\boldsymbol{\theta})}
\newcommand{\Pmath}{\boldsymbol{P}(\boldsymbol{\theta})}
\newcommand{\eye}{\boldsymbol{I}}

\usepackage[bottom]{footmisc}

\title{Isotropic SGD: a Practical Approach to Bayesian Posterior Sampling}

\author{%
  Giulio Franzese\\
  Politecnico di Torino, Italy\\
  EURECOM, Biot, France\\
  \And
  Rosa Candela\\ 
  EURECOM, Biot, France\\
  \And
  Dimitrios Milios\\ 
  EURECOM, Biot, France\\
  \And
  Maurizio Filippone\\ 
  EURECOM, Biot, France\\
  \And
  Pietro Michiardi\\ 
  EURECOM, Biot, France\\
}
\begin{document}

\maketitle

\begin{abstract}
In this work we define a unified mathematical framework to deepen our understanding of the role of stochastic gradient (\sg) noise on the behavior of Markov chain Monte Carlo sampling (\sgmcmc) algorithms. 

Our formulation unlocks the design of a novel, practical approach to posterior sampling, which makes the \sg noise isotropic using a fixed learning rate that we determine analytically, and that requires weaker assumptions than existing algorithms. In contrast, the common traits of existing \sgmcmc algorithms is to approximate the isotropy condition either by drowning the gradients in additive noise (annealing the learning rate) or by making restrictive assumptions on the \sg noise covariance and the geometry of the loss landscape.

Extensive experimental validations indicate that our proposal is competitive with the state-of-the-art on \sgmcmc, while being much more practical to use.

% \giulio{Too Long!}\pietro{Shorter version is in}
% \pietro{Shall we change the title?}
% \pietro{I have modified the template as this was a mix between icml and nips.}
% \pietro{Giulio, could you please modify the citation style? It should use the cite command, not the citep. I will change some of them while making my pass, but please double check everything for coherence.}
\end{abstract}

\section{Introduction}
\label{sec:introduction}

Despite mathematical elegance and some promising results restricted to simple models, standard stochastic gradient (\sg) methods for \mcmc-based Bayesian posterior sampling~\cite{welling2011bayesian,ahn2012bayesian,NIPS2013_4883,chen2014stochastic,ma2015complete} fall short in dealing with the complexity of the loss landscape of deep models~\cite{draxler2018essentially, garipov2018loss}, for which stochastic optimization poses serious challenges~\cite{chaudhari2018stochastic, maddox2019simple}. Moreover, existing methods are often unpractical, as they require ad-hoc, sophisticated vanishing learning rate schedules, and hyper-parameter tuning. 

In general, \sgmcmc algorithms inject random noise to \sg descent algorithms: the covariance of such noise and the learning rate are tightly related to the assumptions on the loss landscape, which together with the \sg noise, determine their sampling properties~\cite{ma2015complete}. 
However, current \sgmcmc algorithms applied to popular complex models such as Deep Nets, cannot satisfy the simplifying assumptions on loss landscapes and on the behavior of the \sg noise covariance, while operating with practical requirements, such as non-vanishing learning rates and ease of use.
A recent work \cite{mandt2017stochastic} argues for fixed step sizes, but settles for variational approximations of simple quadratic losses.
In a generic Bayesian deep learning setting, none of the existing implementations of \sgmcmc methods converge to the true posterior without learning rate annealing.

While we are not the first to highlight these issues ~\cite{draxler2018essentially, garipov2018loss}, including the lack of a unified notation \citep{ma2015complete}, we believe that studying the role of noise in \sgmcmc algorithms has not received enough attention, and a deeper understanding is truly desirable, as it can clarify how various methods compare. 
Most importantly, this endeavor can suggest novel and more practical algorithms relying on fewer tunable parameters and less restrictive assumptions.

In this work, we present a mathematical framework that emphasizes the role of \sg noise covariance and learning rate on the behavior of \sgmcmc algorithms (Sec.~\ref{sec:backandrel}). As a result, the equivalence between learning rate annealing and the injection of noise with extremely large variance becomes clear, and this allows us to propose a novel, practical \sgmcmc algorithm (Sec.~\ref{sec:methodes}) that produces (approximate) posterior samples with a fixed, easy to derive, learning rate. Furthermore, our approach can be readily applied to pre-trained models: after a ``warm-up'' phase to compute \sg noise estimates, it can efficiently perform Bayesian posterior sampling. The proposed \sgmcmc method is a valid, theoretically sound, and simple alternative to popular techniques, that have shortcomings when it comes to the assumptions they rely on \cite{gal2016dropout}.

% We derive our proposal, by first analyzing the case where we inject the smallest complementary noise such that its combined effects with the \sg noise result in an isotropic noise. 
% Thanks to this isotropic property of the noise, it is possible to deal with intricate loss surfaces typical of  deep models, and produce samples from the true posterior without learning rate annealing. 
% This, however, comes at the expense of cubic complexity matrix operations.
% We address such issues through a practical variant of our scheme, which employs well-known approximations to the \sg noise covariance (see, e.g., \citep{springenberg2016bayesian}).
% The result is an algorithm that produces approximate posterior samples with a fixed, easy to derive, learning rate. 
% Note that in generic Bayesian deep learning setting, none of the existing implementations of \sgmcmc methods converge to the true posterior without learning rate annealing. In contrast, our method automatically determines an appropriate learning rate through a simple estimation procedure. Furthermore, our approach can be readily applied to pre-trained models: after a ``warm-up'' phase to compute \sg noise estimates, it can efficiently perform Bayesian posterior sampling.

We evaluate \sgmcmc algorithms (Sec. \ref{sec:experiments}) through an extensive experimental campaign, where we compare our approach to a number of alternatives, including Monte Carlo Dropout (\mcd) \citep{gal2016dropout} and Stochastic Weighted Averaging Gaussians (\swag, \citep{maddox2019simple}), which have been successfully applied to the Bayesian deep learning setting. 
Our results indicate that our approach is simpler to use than other families of \sgmcmc methods, and offer performance that are competitive to the state-of-the-art, in terms of accuracy and uncertainty.

\section{Preliminaries and Related Work}
\label{sec:backandrel}
% Stochastic gradient (\sg) methods have been extensively studied as a means for \mcmc-based Bayesian posterior sampling algorithms to scale to large data regimes. Variants of \sgmcmc algorithms have been studied through the lenses of first \cite{welling2011bayesian,ahn2012bayesian,NIPS2013_4883} or second order \cite{chen2014stochastic,ma2015complete} Langevin Dynamics, which are mathematically convenient continuous-time processes that correspond to discrete-time gradient methods with and without momentum, respectively.
% The common traits underlying many methods from the literature can be summarized as follows: they address large data requirements using \sg and mini-batching, they inject Gaussian noise throughout the algorithm execution, and they avoid the expensive Metropolis-Hasting accept/reject tests that use the whole data~\cite{welling2011bayesian, ahn2012bayesian, chen2014stochastic}.

Consider a data-set of $m-$dimensional observations $\mathcal{D}=\{\boldsymbol{U}_i\}_{i=1}^N$ and a statistical model defined through a likelihood function $p(\mathcal{D}|\thetavect)$ parameterized through a $d$-dimensional set of parameters $\thetavect$. Given a prior $p(\thetavect)$ the posterior over the parameters is obtained by means of Bayes theorem as:% as follows:

\begin{equation}
\label{eq:bayes}
p(\thetavect|\mathcal{D}) = \frac{p(\mathcal{D}|\thetavect) \, p(\thetavect)}{p(\mathcal{D})},
\end{equation}
where $p(\mathcal{D})$ is also known as the model evidence, defined as the integral $p(\mathcal{D}) = \int p(\mathcal{D}|\thetavect) \, p(\thetavect) d\thetavect$.
% Except when the prior and the likelihood function are conjugate, Eq.~\eqref{eq:bayes} is analytically intractable \cite{Bishop06}.
% However, the joint likelihood term in the numerator is typically not hard to compute;
% this is a key element of many \mcmc algorithms, since the normalisation constant $p(\mathcal{D})$ does not affect the shape of the distribution in any way other than scaling.
The posterior distribution is necessary in order to obtain predictive distributions for new test observations $\boldsymbol{U}_*$: $p(\boldsymbol{U}_* | \mathcal{D}) = \int p(\boldsymbol{U}_* | \thetavect) p(\thetavect | \mathcal{D}) d\thetavect$.
% \begin{equation}
% \label{eq:predictive}
% p(\boldsymbol{U}_* | \mathcal{D}) = \int p(\boldsymbol{U}_* | \thetavect) p(\thetavect | \mathcal{D}) d\thetavect.
% \end{equation}
We focus on Monte Carlo methods to obtain an estimate of this predictive distribution, by averaging over $N_{\mathrm{MC}}$ samples obtained from the posterior over $\thetavect$, that is $\thetavect^{(i)} \sim p(\thetavect | \mathcal{D})$:
\begin{equation}
\label{eq:predictive:montecarlo}
p(\boldsymbol{U}_* | \mathcal{D}) \approx \frac{1}{N_{\mathrm{MC}}} \sum_{i=1}^{N_{\mathrm{MC}}} p(\boldsymbol{U}_* | \thetavect^{(i)}). 
\end{equation}

Since Eq.~\eqref{eq:bayes} is analytically intractable \cite{Bishop06}, unless the prior is conjugate to the likelihood, we work with an unnormalized version of the logarithm of the posterior density, and express the negative logarithm of the joint distribution of the data-set $\mathcal{D}$ and parameters $\thetavect$ as:
\begin{equation}
   -f(\thetavect)=\sum\limits_{i=1}^N \log p(\boldsymbol{U}_i | \thetavect)+\log p(\thetavect).
\end{equation}

For computational efficiency, we use a mini-batch stochastic gradient $\boldsymbol{g}(\thetavect)$, which guarantees that the estimated gradient is an unbiased estimate of the true gradient $\nabla f(\thetavect)$, and we assume that the randomness due to the mini-batch introduces a Gaussian noise:\footnote{This is an ordinary assumption in the literature. In light of recent work \cite{csimcsekli2019heavy,pmlr-v97-simsekli19a}, we will address the case for an $\alpha$-stable \sg noise distribution in \Cref{sec:methodes}.} $\boldsymbol{g}(\thetavect)\sim N( \nabla f(\thetavect), 2 \Bmath)$,
% \begin{equation}\label{gaussmb}
% \boldsymbol{g}(\thetavect)\sim N( \nabla f(\thetavect), 2 \Bmath),
% \end{equation}
where the matrix $\Bmath$ denotes the \sg noise covariance, which depends on the parametric model, the data distribution and the mini-batch size.

A survey of algorithms to sample from the posterior using \sg methods can be found in \cite{ma2015complete}. In \Cref{appa} we complement this section with a full derivation, well-known facts and definitions, for completeness. 
As shown in the literature \cite{chen2016bridging, mandt2017stochastic}, there are structural similarities between \sgmcmc algorithms and stochastic optimization methods, and both can be used to draw samples from posterior distributions. In what follows, we define a unified notation to compare many existing algorithms in light of the role played by their noise components.

Stochastic gradient descent (\sgd) can be studied through the following stochastic differential equation (\sde) \cite{gardiner2004handbook, kushner2003stochastic, Ljung1992stochastic}, when the learning rate $\eta$ is small enough:
\begin{equation}\label{eq:sde0}
d\zetavect_t=\boldsymbol{s}(\zetavect_t)dt+\sqrt{2 \eta \boldsymbol{D}(\zetavect_t)}d\boldsymbol{W}_t.
\end{equation}
Here we use a generic form of the \sde, with variable $\zetavect$ instead of $\thetavect$, that accommodates \sgd variants, with and without momentum. 
It is typical \sgmcmc practice \cite{welling2011bayesian,ahn2012bayesian,NIPS2013_4883,mandt2017stochastic} to allow the stochastic process induced by Eq.~\eqref{eq:sde0} to go through an initial adaptation phase where the learning rate is annealed, followed by fixing the learning rate to a small value, to ensure the process reaches and maintains a stationary distribution thereafter.

\begin{definition}
A distribution $\rho(\zetavect)\propto\exp(-\phi(\zetavect))$ is said to be a \textbf{stationary} distribution for the \sde of the form \eqref{eq:sde0}, if and only if it satisfies the following Fokker-Planck equation (\fpe): 
\begin{flalign}
0=\mathrm{Tr}\left\{
    \nabla \left[
        -\boldsymbol{s}(\zetavect)^\top \rho(\zetavect)
        +\eta \nabla^\top \left( 
            \boldsymbol{D} \left( \zetavect \right) \rho(\zetavect)
            \right)
        \right]
    \right\}.
\end{flalign}
\end{definition}
Note that, the operator $\nabla^\top$ applied to matrix $\boldsymbol{D}(\zetavect)$ produces a row vector whose elements are the divergences of the $\boldsymbol{D}(\zetavect)$ columns \cite{chen2014stochastic}.

In general, the stationary distribution does not converge to the desired posterior distribution, i.e., $\phi(\zetavect)\neq f(\zetavect)$, as shown by \cite{chaudhari2018stochastic}. Additionally, given an initial condition for $\zetavect_t$, its distribution is going to converge to $\rho(\zetavect)$ only for $t\rightarrow \infty$. 

Next, we revisit known approaches to Bayesian posterior sampling, and interpret them as variants of an \sgd process, using the \fpe formalism. In what follows, we use $n$ to indicate discrete time, and $t$ for continuous time.

% \subsection{Gradient methods without momentum}\label{sec:1stord}
\noindent \textbf{Gradient methods without momentum.} The generalized update rule of \sgd, described as a discrete-time stochastic process, writes as: $\boldsymbol{\delta} \thetavect_{n}=-\eta \boldsymbol{P}(\thetavect_{n-1})( \boldsymbol{g}(\thetavect_{n-1})+\boldsymbol{w}_{n})$,
% \begin{flalign}
% \boldsymbol{\delta} \thetavect_{n}=-\eta \boldsymbol{P}(\thetavect_{n-1})( \boldsymbol{g}(\thetavect_{n-1})+\boldsymbol{w}_{n}),
% \end{flalign}
where $\boldsymbol{P}(\thetavect_{n-1})$ is a user-defined preconditioning matrix, and $\boldsymbol{w}_{n}$ is a noise term, distributed as $\boldsymbol{w}_{n}\sim N(\zerovect,2\boldsymbol{C}(\thetavect_{n}))$, with a user-defined covariance matrix $\boldsymbol{C}(\thetavect_{n})$. Then, the corresponding continuous-time \sde is \cite{gardiner2004handbook}: 
\begin{equation}\label{contsgd}
d\thetavect_t=-\boldsymbol{P}(\thetavect_t)\nabla f(\thetavect_t) dt+\sqrt{2\eta \boldsymbol{P}(\thetavect_t)^2\boldsymbol{\Sigma}(\thetavect_t)}d\boldsymbol{W}_t.
\end{equation}
We denote by $\Cmath$ the covariance of the \emph{injected noise} and by  $\boldsymbol{\Sigma}(\thetavect_t)=\boldsymbol{B}(\thetavect_t)+\boldsymbol{C}(\thetavect_t)$ the \emph{composite noise} covariance, which combines the \sg and the injected noise.

We define the stationary distribution of the \sde in \Cref{contsgd} as $ \rho(\thetavect)\propto\exp(-\phi(\thetavect))$, noting that when $\boldsymbol{C}=\zerovect$, the potential $\phi(\thetavect)$ differs from the desired posterior $f(\thetavect)$ \cite{chaudhari2018stochastic}. The following theorem, which is an adaptation of known results in light of our formalism, states the conditions for which the \emph{noisy} \sgd converges to the true posterior distribution (proof in \cref{sec:1stord}).

\begin{theorem}\label{theo1}
Consider dynamics of the form \eqref{contsgd} and define the stationary distribution $ \rho(\thetavect)\propto\exp(-\phi(\thetavect))$. If
% \begin{equation}\label{cond1}
%     \nabla^\top\left(\Sigmamath^{-1}\right)=\zerovect^\top\,\,\,\,\text{and}\,\,\,\, \eta \Pmath=\Sigmamath^{-1},
% \end{equation}
$\nabla^\top\left(\Sigmamath^{-1}\right)=\zerovect^\top$ and $\eta \Pmath=\Sigmamath^{-1}$, 
then $\phi(\thetavect)=f(\thetavect)$.
\end{theorem}

\sgld \cite{welling2011bayesian} is a simple approach to satisfy \Cref{theo1}; it uses no preconditioning, $\Pmath=\eye$, and sets the injected noise covariance to $\Cmath=\eta^{-1}\eye$. In the limit for $\eta \rightarrow 0$, it holds that $\Sigmamath=\Bmath+\eta^{-1}\eye\simeq \eta^{-1}\eye$. Then, $\nabla^\top\left(\Sigmamath^{-1}\right)=\eta\nabla^\top\eye=\zerovect^\top$, and $\eta \Pmath=\Sigmamath^{-1}$.
While \sgld succeeds in (asymptotically) generating samples from the true posterior, its mixing rate is unnecessarily slow, due to the extremely small learning rate \cite{ahn2012bayesian}.

An extension to \sgld is Stochastic Gradient Fisher Scoring (\sgfs) \cite{ahn2012bayesian}, which can be tuned to switch between sampling from an approximate posterior, using a non-vanishing learning rate, and the true posterior, by annealing the learning rate to zero. \sgfs uses  preconditioning, $\Pmath\propto \Bmath^{-1}$. Generally, however, $\Bmath$ is ill conditioned: many of its eigenvalues are almost zero \cite{chaudhari2018stochastic}, and computing $\Bmath^{-1}$ is problematic. Moreover, using a non-vanishing learning rate, the conditions of \Cref{theo1} are met only if, at convergence, $\nabla^\top(\Bmath^{-1})=\zerovect^{\top}$, which would be trivially true if $\Bmath$ was constant. However, recent work \cite{draxler2018essentially,garipov2018loss} suggests that this condition is difficult to justify.

The Stochastic Gradient Riemannian Langevin Dynamics (\sgrld) algorithm \cite{NIPS2013_4883} extends \sgfs to the setting in which $\nabla^\top(\Bmath^{-1}) \neq \zerovect^{\top}$. The process dynamics are adjusted by adding the term $\nabla^\top(\Bmath^{-1})$ which, however, cannot be easily estimated, restricting \sgrld to cases where it can be computed analytically.

The work in \cite{li2016preconditioned} considers a regularized diagonal preconditioning matrix $\Pmath$, derived from the \sg noise. However, in the sampling phase, the method neglects the term $\nabla^\top \Pmath$ and the regularization term is a user-defined parameter, which is not trivial to tune.

The approach in \cite{mandt2017stochastic} investigates constant-rate \sgd (with no injected noise), and determines analytically the learning rate and preconditioning that minimize the Kullback–Leibler (\name{kl}) divergence between an approximation and the true posterior. Moreover, it shows that the preconditioning used in \sgfs is optimal, in the sense that it converges to the true posterior, when $\Bmath$ is constant and the true posterior has a quadratic form. 

% \pietro{Also say that they use tempered distributions?}
In summary, to claim convergence to the true posterior distribution, existing approaches require either vanishing learning rates or assumptions on the \sg noise covariance that are difficult to verify in practice, especially when considering deep models. We instead propose a novel practical method, that induces isotropic \sg noise and thus satisfies \Cref{theo1}. We determine analytically a fixed learning rate and we require weaker assumptions on the loss geometry.

% \subsection{Gradient methods with momentum}
\noindent \textbf{Gradient methods with momentum.} Momentum-corrected methods emerge as a natural extension of \sgd approaches. The general set of update equations for (discrete-time) momentum-based algorithms is:
\begin{flalign*}
\begin{cases}
\boldsymbol{\delta} \thetavect_{n}=\eta \boldsymbol{P}(\thetavect_{n-1})\boldsymbol{M}^{-1} \boldsymbol{r}_{n-1}\\
\boldsymbol{\delta} \boldsymbol{r}_{n}=-\eta \boldsymbol{A}(\thetavect_{n-1}) \boldsymbol{M}^{-1}\boldsymbol{r}_{n-1}-\eta \boldsymbol{P}(\thetavect_{n-1})( \boldsymbol{g}(\thetavect_{n-1})+\boldsymbol{w}_{n}),
\end{cases}
\end{flalign*}

where $\boldsymbol{P}(\thetavect_{n-1})$ is a preconditioning matrix, $\boldsymbol{M}$ is the mass matrix and $\boldsymbol{A}(\thetavect_{n-1})$ is the friction matrix~\cite{chen2014stochastic, neal2011mcmc}. Similarly to the first order counterpart, the noise term is distributed as $\boldsymbol{w}_{n}\sim N(\zerovect,2\boldsymbol{C}(\thetavect_{n})))$. Then, the \sde to describe continuous-time system dynamics is:
\begin{flalign}\label{sysmom}
\begin{cases}
d\thetavect_t= \boldsymbol{P}(\thetavect_t)\boldsymbol{M}^{-1} \boldsymbol{r}_t dt\\
d\boldsymbol{r}_t=- (A(\thetavect_t) \boldsymbol{M}^{-1}r_t+\boldsymbol{P}(\thetavect_t)\nabla f(\thetavect_t))dt+ \sqrt{2\eta \boldsymbol{P}(\thetavect_t)^2\boldsymbol{\Sigma}(\thetavect_t)}d\boldsymbol{W}_t.
\end{cases}
\end{flalign}
where $\boldsymbol{P}(\thetavect_t)^2=\boldsymbol{P}(\thetavect_t)\boldsymbol{P}(\thetavect_t)$, and $\boldsymbol{P}(\thetavect_t)$ is symmetric. 
The theorem hereafter (proof in \Cref{sec:2ndord}) describes the conditions for which the \sde converges to the true posterior distribution.
\begin{theorem}\label{theo2}
Consider dynamics of the form \eqref{sysmom} and define the stationary distribution for $\thetavect_t$ as $ \rho(\thetavect)\propto\exp(-\phi(\thetavect))$. If
% \begin{equation}\label{cond2}
% \nabla^\top \Pmath=\zerovect^{\top}\,\,\,\,\text{and}\,\,\,\, \Amath=\eta \Pmath^2\Sigmamath,
% \end{equation}
$\nabla^\top \Pmath=\zerovect^{\top}$ and $\Amath=\eta \Pmath^2\Sigmamath$, 
then $\phi(\thetavect)=f(\thetavect)$ .
\end{theorem}
In the naive case, where $\Pmath=\eye,\Amath=\zerovect,\Cmath=\zerovect$, the conditions in \Cref{theo2} are not satisfied and the stationary distribution does not correspond to the true posterior \cite{chen2014stochastic}. To generate samples from the true posterior, it is sufficient to set $\Pmath=\eye,\Amath=\eta \Bmath,\Cmath=\zerovect$ (as in Eq.~(9) in \cite{chen2014stochastic}). 

Stochastic Gradient Hamiltonian Monte Carlo (\sghmc) \cite{chen2014stochastic} suggests that estimating $\Bmath$ can be costly. Hence, the injected noise $\Cmath$ is chosen such that $\Cmath=\eta^{-1} \Amath$, where $\Amath$ is user-defined. When $\eta \rightarrow 0$, the following approximation holds: $\Sigmamath\simeq \Cmath$. It is then trivial to check that conditions in \Cref{theo2} hold without the need for explicitly estimating $\Bmath$. A further practical reason to avoid setting $\Amath=\eta \Bmath$ is that the computational cost for the operation $\boldsymbol{A}(\thetavect_{n-1}) \boldsymbol{M}^{-1}\boldsymbol{r}_{n-1}$ has $\mathcal{O}(D^2)$ complexity, whereas if $\Cmath$ is diagonal, this is reduced to $\mathcal{O}(D)$. This however, severely slows down the sampling process.

Stochastic Gradient Riemannian Hamiltonian Monte Carlo (\sgrhmc) is an extension to \sghmc \cite{ma2015complete}), which considers a generic, space-varying preconditioning matrix $\Pmath$ derived from information geometric arguments \cite{girolami2011riemann}.  \sgrhmc suggests to set $\Pmath=\boldsymbol{G}(\thetavect)^{-\frac{1}{2}}$, where $\boldsymbol{G}(\thetavect)$ is the Fisher Information matrix. To meet the requirement $\nabla^\top \Pmath=\zerovect^\top$, it includes a correction term, $-\nabla^\top \Pmath$. The injected noise is set to  $\Cmath=\eta^{-1}\eye-\Bmath$, consequently $\boldsymbol{\Sigma}=\eta^{-1}\eye$, and the friction matrix is set to  $\Amath=\Pmath^{2}$. With all these choices, \Cref{theo2} is satisfied. While appealing, the main drawbacks of this method are the need for an analytical expression of $\nabla^\top \Pmath$, and the assumption for $\Bmath$ to be known.

Recently, the work in \cite{Zhang2020Cyclical} suggests to use a cyclical learning rate schedule to better explore the loss landscape and sample more efficiently. While the idea is appealing, it introduces further hyper-parameters to tune, which is opposite to our quest for a simple, easy to use method. 
% Furthermore, similarly to most work in the literature, it uses temperature scaling to help with the convergence, which harms the process to sample from the true posterior.

From a practical standpoint, momentum-based methods suffer from the complexity and fragility of hyper-parameter tuning, including the learning rate schedule and those that govern the simulation of a second-order Langevin dynamics. 
The method we propose in this work can be applied to momentum-based algorithms as well; then, it can be viewed as an extension to the work in \cite{springenberg2016bayesian}, albeit addressing more complex loss landscapes. However, we leave this avenue of research for future work.

\section{Sampling by layer-wise Isotropization}
\label{sec:methodes}
We present a simple and practical approach to inject noise to SGD iterates to perform Bayesian posterior sampling.
Our goal is to sample from the true posterior distribution (or approximations thereof) using a \emph{constant} learning rate, and to rely on more lenient assumptions about the geometry of the loss landscape that characterize deep models, compared to previous works.

From Theorem \ref{theo1}, observe that $\Pmath,\Sigmamath$ are instrumental to determine the convergence properties of \sg methods to the true posterior. 
We consider the constructive approach of \textit{designing} $\eta \Pmath$ to be a constant, diagonal matrix, constrained to be layer-wise uniform:
\begin{equation}\label{eqdiagcov}
    \eta \Pmath= \Lambdamath^{-1}=\mathrm{diag} ( [ \underbrace{\lambda^{(1)},\dots,\lambda^{(1)}}_{\text{layer 1}},\dots, \underbrace{\lambda^{(N_l)},\dots\lambda^{(N_l)}}_{\text{layer $N_l$}} ] )^{-1}.
\end{equation}
By properly setting parameters $\lambda^{(p)}$, we achieve the simultaneous result of a non-vanishing learning rate and a well-conditioned preconditioning matrix. This implies a layer-wise learning rate  $\eta^{(p)}=\frac{1}{\lambda^{(p)}}$ for the $p$-{th} layer, without further preconditioning.
We can now state (see proof in \Cref{cor1proof}), as a corollary to Theorem \ref{theo1}, that our method guarantees convergence to the true posterior distribution.

\begin{corollary}\label{theo_isgd} %(Theorem \ref{theo1}) 
Given the dynamics of Eq.~\eqref{contsgd} and the stationary distribution $\rho(\thetavect)\propto\exp(-\phi(\thetavect))$, if $\eta \Pmath= \Lambdamath^{-1}$ as in Eq.~\eqref{eqdiagcov}, and $\Cmath=\Lambdamath-\Bmath \succ 0\, \forall \thetavect$, then $\phi(\thetavect)=f(\thetavect)$.
\end{corollary}
If the above conditions hold, it is simple to show that matrices $\Pmath$ and $\Cmath$ satisfy \Cref{theo1}.
Then, we say that the composite noise covariance $\Sigmamath = \Cmath+\Bmath = \mathrm{diag}\left(\begin{bmatrix}\lambda^{(1)},\dots,\lambda^{(1)}, \dots, \lambda^{(N_l)},\dots\lambda^{(N_l)}\end{bmatrix}\right)$ is \emph{isotropic} within model layers. We set $\Lambdamath$ to be, among all valid matrices satisfying $\Lambdamath - \Bmath\succ 0$, the smallest, i.e., the one  with the smallest $\lambda$'s. Indeed, larger $\Lambdamath$ induces a small learning rate, thus unnecessarily reducing sampling speed. 

\noindent \textbf{An ideal method.} Now, let's consider an ideal case, in which we assume the \sg noise covariance $\Bmath$ and $\Lambdamath$ to be known in advance.
The procedure described in Algorithm \ref{oraclealgo} illustrates a naive \sg method that uses the \emph{injected noise} covariance $\Cmath$ to sample from the true posterior.

\begin{minipage}{0.46\textwidth}
\begin{algorithm}[H]
\caption{Idealized posterior sampling}\label{oraclealgo}
\begin{algorithmic}
      \State \textbf{\textsc{Sample} ($\thetavect_{0}, \Bmath, \Lambdamath$):}
      \State $\thetavect \leftarrow \thetavect_0$ \Comment{Initialize $\thetavect_t$}
      \Loop
      \State $\boldsymbol{g}=\frac{\nabla \tilde{f}(\thetavect)}{N}$ \Comment{Compute \sg}
      \State $\Cmath^{\frac{1}{2}}\leftarrow ( \boldsymbol{\Sigma}-\Bmath)^{\frac{1}{2}}$\label{sqrtoper}
      \State $\boldsymbol{n}\sim N(0,\eye)$
      \State $\boldsymbol{w}\leftarrow \Cmath^{\frac{1}{2}}\boldsymbol{n}$
      \State $\delta\thetavect \leftarrow (N\boldsymbol{\Sigma})^{-1}(\boldsymbol{g}+\sqrt{2}\boldsymbol{w})$
      \State $\thetavect \leftarrow \thetavect-\delta\thetavect$ \Comment{Update $\thetavect$}
    \EndLoop
\end{algorithmic}
\end{algorithm}
\end{minipage}
\begin{minipage}{0.46\textwidth}
\begin{algorithm}[H]
   \caption{\isgd: practical sampling}\label{samplealgo}
    \begin{algorithmic}
      \State \textbf{\textsc{Sample}} ($\thetavect_{0}$):
      \State $\thetavect\leftarrow\thetavect_{0}$ \Comment {Initialize $\thetavect_t$}
      \Loop
      \State $\boldsymbol{g}=\frac{\nabla \tilde{f}(\thetavect)}{N}$ \Comment{Compute \sg}
    %   \For{$p \leftarrow 1$ to $N_l$} 
       \State $\Cmath^{\frac{1}{2}(p)}\leftarrow (\lambda^{(p)}-\frac{1}{2}(\tilde{\boldsymbol{\sigma}}(\thetavect)^{(p)})^{2})^{\frac{1}{2}}$
      \State $\boldsymbol{n}\sim N(\zerovect,\eye)$
            \State $\boldsymbol{w}\leftarrow \Cmath^{\frac{1}{2}(p)}\boldsymbol{n}$
      \State $\delta \thetavect^{(p)} \leftarrow (N\lambda^{(p)})^{-1} \left( \boldsymbol{g}^{(p)} + \sqrt{2}\boldsymbol{w}\right)$
    %   \EndFor
      \State $\thetavect \leftarrow \thetavect - \delta \thetavect$\Comment{Update $\thetavect$}
     \EndLoop
\end{algorithmic}
\end{algorithm}
\end{minipage}

This deceivingly simple procedure is guaranteed to generate samples from the true posterior, with a non-vanishing learning rate\footnote{Note that instead of computing the gradient of $f(\thetavect)$, we compute the (mini-batch) gradient of $\frac{f(\thetavect)}{N}$, similarly to the notation used in \cite{mandt2017stochastic}, that we indicate with $\frac{\nabla \tilde{f}(\thetavect_t)}{N}$.}. 
However, it cannot be used in practice as $\Bmath$ and $\Lambdamath$ are unknown. Also, the algorithm is costly, as it requires computing $( \boldsymbol{\Sigma}-\Bmath)^{\frac{1}{2}}$, which requires $\mathcal{O}(d^{3})$ operations, and $\Cmath^{\frac{1}{2}}$, which costs $\mathcal{O}(d^2)$ multiplications.
Next, we describe a practical approach, where we use approximations at the expense of generating samples from the true posterior distribution. Note that \cite{mandt2017stochastic} suggests to explore a related preconditioning, but does not develop the idea.

\noindent \textbf{A practical method: Isotropic SGD.} To make the idealized sampling method practical, we require additional assumptions which are milder than what is required by current approaches in the literature, as we explain at the end of this section.

%\giulio{ I don't like talking about covariances, it makes no sense to me. Should we rephrase the assumptions using the word scale? Also have a look at pseudocode! (estim of covariance)}

\begin{assumption}\label{ass1}
%The noise components are assumed to be independent.
The \sg noise covariance $\Bmath$ can be approximated with a diagonal matrix, i.e., $\Bmath=\mathrm{diag}(\boldsymbol{b}(\thetavect))$. Thus, the noise components are independent \cite{springenberg2016bayesian,ahn2012bayesian}.
\end{assumption}

\begin{assumption}\label{ass2}
%The signal to noise ratio (SNR) of a gradient is small enough such that, in the stationary regime, for the purpose of noise parameters estimation, using the \sg instead of its mean centered version induces negligible errors.
The signal to noise ratio (SNR) of a gradient is small enough such that, in the stationary regime, the un-centered variance of the gradient is a good estimate of the true variance \cite{springenberg2016bayesian,saxe2019information}. Hence, combining with assumption \ref{ass1}, $\boldsymbol{b}(\thetavect)\simeq \frac{\mathrm{E} [\boldsymbol{g}(\thetavect) \odot \boldsymbol{g}(\thetavect)]}{2}$.
\end{assumption}

\begin{assumption}\label{ass3}
%There exist constants $\beta^{(p)}$, layer by layer, that bound the maximum scale of the noise for the parameters belonging to $p_{th}$ layer, i.e. $1$
In the stationary regime, the maximum of the variances of noise components, layer by layer, are fixed constants (similarly to \cite{zhu2018anisotropic}): $\beta^{(p)} =\max_{j\in I_p} \mathbf{b}_j(\thetavect)$, where $I_p$ is the set of indexes of parameters belonging to $p_{th}$ layer.
\end{assumption}

Note that \Cref{ass2} and \Cref{ass3} must hold only in the stationary regime when the process reaches the bottom valley of the loss landscape. 
Given our assumptions, and our design choices, it is then possible to show (see \cref{lamopt}) that the optimal (i.e., the smallest possible) $\Lambdamath = \begin{bmatrix}\lambda^{(1)},\dots,\lambda^{(1)}, \dots, \lambda^{(N_l)},\dots\lambda^{(N_l)}\end{bmatrix}$ satisfying Corollary \ref{theo_isgd} can be obtained as $\lambda^{(p)} = \beta^{(p)}$. 

Since we cannot assume $\Bmath$ to be known, in what follows we discuss two approaches to estimate its components.
A simple method to estimate $\Bmath$ is as follows (see \Cref{lamest}): we compute $\lambda^{(p)}=\max_{j\in I_p}\mathbf{b}_j(\thetavect)=\frac{1}{2}\max_{j}(\boldsymbol{g}_j(\thetavect)^{(p)})^2$,%$\lambda^{(p)}=\sum_{j\in I_p}b_j(\thetavect)=\frac{1}{2}||\boldsymbol{g}^{(p)}(\thetavect)||^2$, 
where $\boldsymbol{g}(\thetavect)^{(p)}$ is the portion of stochastic gradient corresponding to the $p$-th layer. Our estimates can be extended to use a moving average approach. Our empirical validation, however, indicates that this simple method does not produce stable estimates.

Indeed, a shared assumption of \sgmcmc methods is that \sg noise is Gaussian. While this assumption can be justified with the C.L.T. for relatively simple models (linear models or simple feed-forward networks), its validity has been challenged in the deep learning domain \cite{csimcsekli2019heavy,pmlr-v97-simsekli19a} (see \Cref{lamest} for a detailed discussion), suggesting that, for complex architectures, the noise distribution is heavy tailed. Then, the hypothesis is that the various components of \sg noise follow an \textit{$\alpha$-stable} distribution: $w\sim p(w)$, where $\int_{-\infty}^{+\infty}\exp\left(j2\pi wt\right)p(w) dw=\exp(-|ct|^{\alpha})$, with $\alpha \in (0,2]$, where $\alpha,c$ can vary across different parameters. When $\alpha=2$, $p(w)$ becomes Gaussian, but for $\alpha<2$, its variance goes to infinity, thus estimating the \sg noise covariance is problematic.

Prior works \cite{csimvsekli2017fractional}, suggest to define a stochastic process governed by an \sde that uses L\'{e}vy Noise instead of a Brownian motion. However, this approach comes with its own challenges, such as the approximation of a fractional derivative, and the use of full gradients.
Instead, we propose the following approximate method: we consider that the \sg noise follows a Gaussian distribution, with parameters set to minimize the $l_2$-distance between $p(w)=\sqrt{2\pi\sigma^2}\exp\left(-\frac{w^2}{2\sigma^2}\right)$ and $q(w)=\int_{-\infty}^{+\infty}\exp\left(-j2\pi wt\right)\exp(-|ct|^{\alpha}) dt$ (see \Cref{lamest} for details). We estimate the parameters of the $\alpha$-stable distribution by extending
\cite{vehel2018explicit} to space varying settings, and derive the equivalent variances $\tilde{\sigma}^2$ that minimize the $l_2$ distance.
Then, the $\lambda^{(p)}$ are computed as $\frac{1}{2}\max_{j}(\tilde{\boldsymbol{\sigma}}_j(\thetavect)^{(p)})^{2}$.  %$\frac{1}{2}\sum_{j\in I_p}\tilde{\sigma}_{j}^2$. 
% Our experiments indicate that this method is more stable than the simple approach discussed above.
In Sec.~\ref{sec:experiments} we use this method, as it is more stable than the simple approach discussed above.

Ultimately, the composite noise matrix $\boldsymbol{\Sigma} = \Lambdamath$ is a layer-wise isotropic covariance matrix, which inspires the name of our proposed method as \textit{Isotropic} SGD (\isgd). Once all parameters $\lambda^{(p)}$ have been estimated, the layer-wise learning rate is determined {\it automatically}: for the $p$-th layer, the learning rate is  $\eta^{(p)}=\frac{1}{N \lambda^{(p)}}$. 

The practical implementation of \isgd is shown in Algorithm~\ref{samplealgo}. 
The computational cost of \isgd is as follows. Similarly to \cite{chen2014stochastic}, we define the cost of computing a gradient mini-batch as $C_{\boldsymbol{g}}(N_b,d)$. 
Then (see \Cref{lamest}), the computational cost for estimating the noise covariance scales as $\mathcal{O}(d)$ logarithm computations. The computational cost of generating random samples with the desired covariance scales as $\mathcal{O}(d)$ square roots and $\mathcal{O}(d)$ multiplications. The overall cost of our method is the sum of the above terms. Note that the cost of estimating the noise covariance does not depend on the mini-batch size. The space complexity of \isgd is the same as \sghmc and \sgfs and variants: it scales as $\mathcal{O}(N_{\mathrm{MC}}d)$, where $N_{\mathrm{MC}}$ is the number of posterior samples. 

In \cref{appd} we report a simple numerical example where the true posterior is analytically available: visual inspection indicates that our method produces a sensible predictive posterior distribution.

\noindent \textbf{Assumptions and convergence to the true posterior.} Our theory shows that the ideal version of \isgd (\Cref{theo_isgd} holds, and $\Bmath$ is known)  converges to the true posterior with a constant learning rate. This is not the case for existing work. Even when $\Bmath$ is assumed to be known, \sgfs requires the correction term $\nabla^\top \Bmath^{-1}=0$. Also, both \sgrld and \sgrhmc require computing $\nabla^\top \Bmath^{-1}$, for which an estimation procedure is elusive. The method in \cite{springenberg2016bayesian} needs a \textit{constant}, diagonal $\Bmath$, a condition that does not necessarily hold for deep models. 

Since $\Bmath$ is estimated, \isgd can only approximate the true posterior. Despite elegant theoretical studies to claim convergence to the true posterior, several recent works suggest to use \textit{temperature scaling} \cite{li2016preconditioned,maddox2019simple,Zhang2020Cyclical}. Then, in practice, such methods sample from $p_{T}(\thetavect|\mathcal{D})=p(\thetavect|\mathcal{D})^{\frac{1}{T}}$, by using  scaled gradients $\frac{1}{T}\nabla f(\thetavect)$, or by scaling the preconditioning matrix $\eta \Pmath$ by a fixed constant. Ultimately, this implies an approximation to the true posterior.

\begin{table*}[t!]
\caption{Results for regression on \uci data-sets. Bold results indicate the best \mnll performance, underlined results indicate the best \rmse performance. $n$: number of samples, $d$: input dimension.}
\label{tab:uci_results}

\centering
\resizebox{\linewidth}{!}{
\begin{tabular}{lllllll}

\hline
                               & \multicolumn{2}{c}{\textbf{\isgd}} & \multicolumn{2}{c}{\sghmc} & \multicolumn{2}{c}{\mcd} \\
\textbf{\textit{Dataset}} ($n$, $d$)        & \rmse             & \mnll            & \rmse             & \mnll            & \rmse             & \mnll            \\
\hline
\textbf{\textsc{BOSTON}} (506, 13) &  \underline{3.26} $\pm$ 1.14               & \textbf{3.46} $\pm$ 1.83                & 3.41 $\pm$ 1.13                &    3.56 $\pm$ 1.80            &  3.32 $\pm$ 1.01                &  5.26 $\pm$ 2.38         \\
\textbf{\textsc{CONCRETE}} (1030, 8) &  5.47 $\pm$ 0.38               & 12.33 $\pm$ 4.38               &  5.27  $\pm$ 0.46              & 11.77    $\pm$ 2.81            &    \underline{5.01}    $\pm$ 0.43          &   \textbf{6.77} $\pm$ 1.80                         \\
\textbf{\textsc{PROTEIN}} (45730, 9) &  4.77 $\pm$ 0.05               & 4.20 $\pm$ 0.05               &  4.55    $\pm$ 0.04            &   3.90    $\pm$ 0.02          &   \underline{4.49}  $\pm$ 0.02              &  \textbf{3.69} $\pm$ 0.01                     \\
\textbf{\textsc{WINE}} (1599, 11) &  \underline{0.63}    $\pm$ 0.04            & 1.03 $\pm$ 0.11               &  0.64  $\pm$ 0.05             &   \textbf{0.98}    $\pm$ 0.11         &   0.64  $\pm$ 0.05             &   1.04    $\pm$ 0.13           \\
\textbf{\textsc{YACHT}} (308, 6) & 0.69  $\pm$ 0.33              & 4.19 $\pm$ 7.24              & \underline{0.49} $\pm$ 0.17               & \textbf{3.53} $\pm$ 8.22              &   0.57  $\pm$ 0.16            &   4.15    $\pm$ 4.98         \\
\hline
\end{tabular}
}
\end{table*}

\begin{figure*}[!t]
    \centering
    \subfloat[\isgd]{%
    \centering
    \includegraphics[width=0.33\linewidth]{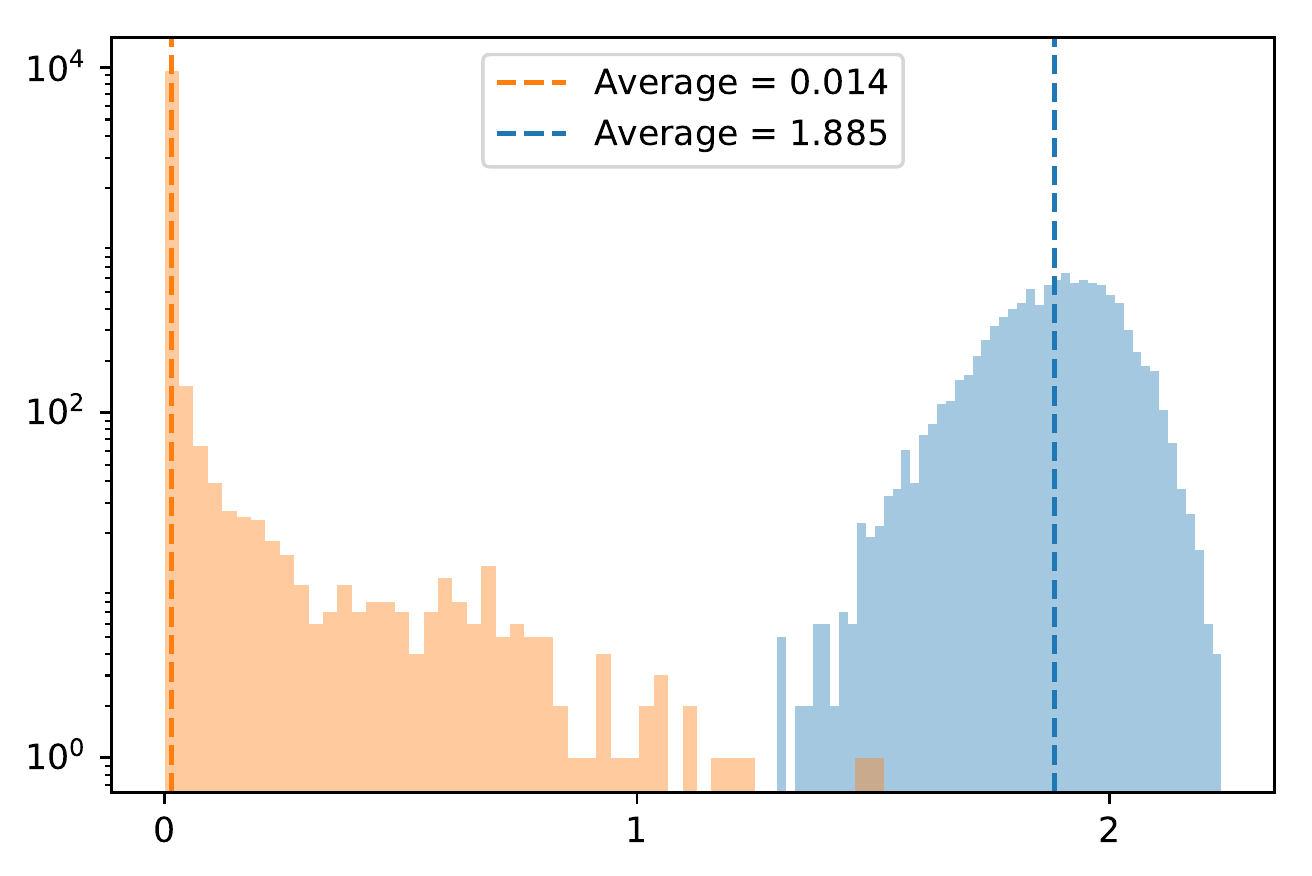}
    }
    \subfloat[\mcd]{%
    \centering
    \includegraphics[width=0.33\linewidth]{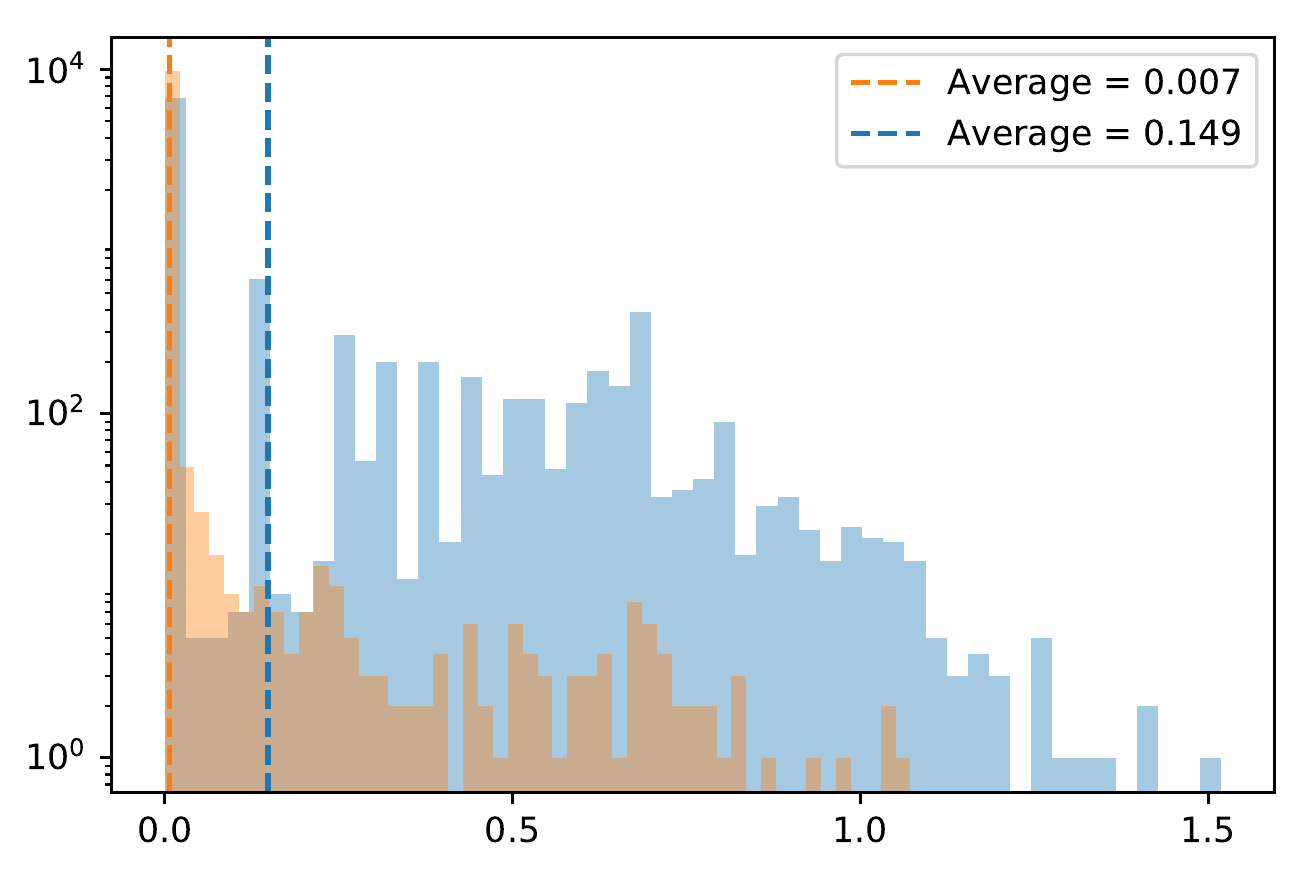}
    }
    \subfloat[\swag]{%
    \centering
    \includegraphics[width=0.33\linewidth]{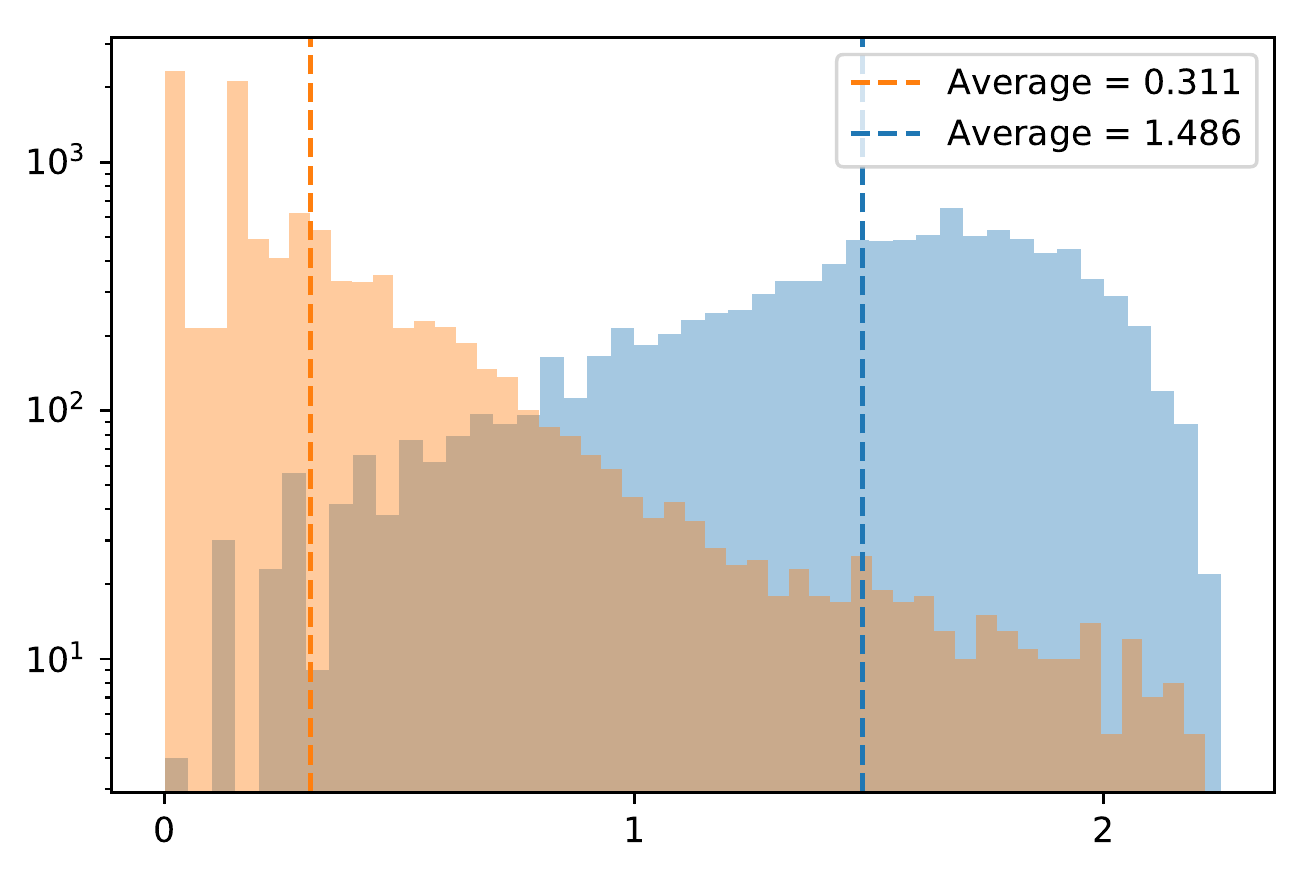}
    }

    \caption{Out-of-distribution entropy histograms on \textsc{mnist} (orange) vs. \textsc{not-mnist} (blue). In this experiment, \isgd outperforms alternative approaches to posterior sampling, including the recent method \swag and the popular method \mcd.}
    \label{fig:entropy}
\end{figure*}

\section{Experiments}
\label{sec:experiments}
We first study \isgd using standard \uci data-sets \cite{Dua:2019} and a shallow feed-forward network. Then, we focus on image classification and use a simple CNN on \mnist \cite{mnist}, and \textsc{ResNet-18}, \textsc{ResNet-56} \cite{he2016deep} and \textsc{VGG-16} \cite{vgg16} on \textsc{Cifar-10} \cite{cifar10}.  We compare \isgd (with the Gaussian approximation to the estimated $\alpha$-stable distribution) to \sghmc \cite{chen2014stochastic}, and to alternative approaches to approximate Bayesian inference, including \mcd \cite{gal2016dropout}, and \swag \cite{maddox2019simple}, the variational \sgd approach (\mandt \cite{mandt2017stochastic}) and \cyc \cite{Zhang2020Cyclical}.
\cref{lamest} includes additional implementation details on \isgd. \cref{appc} presents detailed configurations of all methods we compare and additional results.
%on different architectures, and additional alternative methods.

\noindent \textbf{\uci regression tasks, with a simple model.} We use a simple feed-forward network with two layers and a \textsc{r}e\textsc{lu} activation function; the hidden layer has 50 units. We use the root mean square error (\rmse) for the predictive performance and the mean negative log-likelihood (\mnll) for uncertainty quantification.
At test time we use 100 samples to estimate the predictive posterior distribution, using Eq.~\eqref{eq:predictive:montecarlo}, for \isgd and \sghmc, and 10\,000 samples for \mcd. All our experiments use 10-splits. We note that the task of tuning our competitors is not trivial.\footnote{In \cite{gal2016dropout}, the same task is evaluated using 20 splits and an intricate tuning procedure. Thus, our results are slightly worse than in their work.} 
In this set of experiments we omit results for \swag, which we keep for more involved scenarios.

Tab.~\ref{tab:uci_results} is an overview of our results: \mcd and \isgd are roughly similar, in terms of \rmse. Instead, results for \mnll are more nuanced: all methods are essentially on par, with \isgd outperforming alternatives in some cases.
Additionally, our experiments confirm the practical viability of \isgd, which does not require tedious and creative parameter tuning during the sampling phase.

\noindent \textbf{Classification tasks, with deeper models.} With the exception of \mcd, to compare methods on a fair ground, we use pre-trained models, such that all approaches produce samples from the same conditions. 
% Alternatively, we can train models from scratch and use a burn in phase to reach the conditions for the sampling phase to begin.
As an alternative, we can use a standard optimizer to train a model from scratch: then, \isgd sampling can be triggered when stationarity conditions are met.

Next, we compare \isgd, \mcd, \swag and \mandt using first the \mnist~dataset, and a simple \textsc{LeNet-5} CNN \cite{lecun1998gradient}. We use a pre-trained \textsc{LeNet-5}, and proceed with a ``warm-up'' phase to estimate the \sg noise covariance (this is needed both in \swag, \mandt and \isgd). Then, the posterior sampling phase can begin. At test time we use 30 samples for all methods. Results are obtained averaging over 5 independent seeds. As commonly done in the literature, we also use temperature scaling, which helps stabilizing algorithm dynamics.
All methods are compared based on the test accuracy (\acc) and the \mnll. Additionally, we compute out-of-distribution entropy diagrams by training on \mnist, and testing on both \mnist~and \textsc{not-mnist}. This is common practice to check whether the entropy of the predictions on \textsc{not-mnist} is higher than the entropy of the predictions on \mnist. 
Tab.~\ref{tab:mnist_results} indicates that all methods are essentially equivalent in terms of accuracy. Instead, \isgd outperforms others in terms of \mnll. Fig.~\ref{fig:entropy} indicates that \isgd achieves a better separation of in-distribution (orange) and out-of-distribution (blue) predictions.

% \begin{table}[!t]
% \begin{minipage}{0.5\textwidth}
%     \resizebox{\textwidth}{!}{
%     \begin{tabular}{ccc}
%         \hline
%         Method & \acc   & \mnll\\
%         \hline
%         \textbf{\isgd} & 99.35 $\pm$ 0.05 & \textbf{0.0226} $\pm$ 0.0010 \\
%         \textbf{\mcd} & \textbf{99.38} $\pm$ 0.02 & 0.0242 $\pm$ 0.0017 \\
%         \textbf{\swag} & 99.14 $\pm$ 0.07 & 0.0291 $\pm$ 0.0012 \\
%         % {} & {} & {}\\
%         \hline
%     \end{tabular}
%     }
%     \caption{Performance comparison of \textsc{LeNet-5} on \mnist}
%     \label{tab:mnist_results}
% \end{minipage}
% \begin{minipage}{0.5\textwidth}
%         \resizebox{\textwidth}{!}{
%         \begin{tabular}{ccc}
%         \hline
%         Method & \acc   & \mnll\\
%         \hline
%         \textbf{\isgd} & 94.07 $\pm$ 0.02 & 0.1949 $\pm$ 0.0046 \\
%         \textbf{\mcd} & \textbf{95.15} $\pm$ 0.10 & 0.1734 $\pm$ 0.0033 \\
%         \textbf{\swag} & 94.90 $\pm$ 0.08 & \textbf{0.1551} $\pm$ 0.0016 \\
%         \textbf{\baseline} & 94.32 $\pm$ 0.21 & 0.2045 $\pm$ 0.0032 \\
%         \hline
%     \end{tabular}
%     }
%     \caption{Performance comparison of \textsc{ResNet-56} on \textsc{cifar10}}
%     \label{tab:cifar10_results}
% \end{minipage}
% \end{table}

\begin{table}
\parbox{.5\linewidth}{
    \caption{Performance of \textsc{LeNet-5} on \mnist}
    \label{tab:mnist_results}
    % \resizebox{\textwidth}{!}{
    \centering
    \begin{tabular}{ccc}
        \hline
        Method & \acc   & \mnll\\
        \hline
        \textbf{\isgd} & \textbf{99.42} $\pm$ 0.03 & \textbf{0.0222} $\pm$ 0.0010 \\
        \textbf{\mcd} & 99.38 $\pm$ 0.02 & 0.0242 $\pm$ 0.0017 \\
        \textbf{\swag} & 99.14 $\pm$ 0.07 & 0.0291 $\pm$ 0.0012 \\
        \textbf{\mandt} & 99.41 $\pm$ 0.03 & 0.0224 $\pm$ 0.0012 \\
        % {} & {} & {}\\
        \hline
    \end{tabular}
    }
\parbox{.5\linewidth}{
\caption{Performance of \textsc{ResNet-56} on \textsc{cifar10}}
    \label{tab:cifar10_results}
    \centering
        % \resizebox{\textwidth}{!}{
        \begin{tabular}{ccc}
        \hline
        Method & \acc   & \mnll\\
        \hline
        \textbf{\isgd} & 94.37 $\pm$ 0.15 & 0.2011 $\pm$ 0.0035 \\
        %\textbf{\isgd} & 94.07 $\pm$ 0.02 & 0.1949 $\pm$ 0.0046 \\
        \textbf{\mcd} & \textbf{95.15} $\pm$ 0.10 & 0.1734 $\pm$ 0.0033 \\
        \textbf{\swag} & 94.90 $\pm$ 0.08 & \textbf{0.1551} $\pm$ 0.0016 \\
        \textbf{\mandt} & 93.82 $\pm$ 0.19 & 0.2664 $\pm$ 0.0100 \\
        %\textbf{\baseline} & 94.32 $\pm$ 0.21 & 0.2045 $\pm$ 0.0032 \\
        \hline
    \end{tabular}
}
    % }
\end{table}

\begin{table}
\parbox{.5\linewidth}{
\caption{Performance of \textsc{VGG-16} on \textsc{cifar10}}
    \label{tab:cifar10_vgg_results}
\centering
    \begin{tabular}{c c c}
        \hline
        Method & \acc   & \mnll\\
        \hline
        \textbf{\isgd} & 92.73 $\pm$ 0.07 & 0.3577 $\pm$ 0.0046 \\
        \textbf{\mcd} & 92.71 $\pm$ 0.12 & 0.2470 $\pm$ 0.0067 \\
        \textbf{\swag} & \textbf{93.66} $\pm$ 0.13 & \textbf{0.1946} $\pm$ 0.0036 \\
        \textbf{\mandt} & 93.02 $\pm$ 0.06 & 0.3313 $\pm$ 0.0062 \\
  %      \textbf{\baseline} & 93.07 $\pm$ 0.10 & 0.3302 $\pm$ 0.0065 \\
        \hline
    \end{tabular}
}
\parbox{.5\linewidth}{
\caption{Performance of \textsc{ResNet-18} on \textsc{cifar10}}
    \label{tab:cifar10_res18_results}
\centering
    \begin{tabular}{c c c}
        \hline
        Method & \acc   & \mnll\\
        \hline
        \textbf{\isgd} & 95.38 $\pm$ 0.12 & 0.1794 $\pm$ 0.0044 \\
        \textbf{\cyc} & \textbf{95.73} $\pm$ 0.03 & N/A \\
  %      \textbf{\baseline} & 93.07 $\pm$ 0.10 & 0.3302 $\pm$ 0.0065 \\
        \hline
        & & \\
        & & \\
    \end{tabular}
}
\end{table}

%\begin{table}[!t]
%    \caption{Performance comparison of \textsc{LeNet-5} on \mnist.}
    %\label{tab:mnist_results}
    % \resizebox{\textwidth}{!}{
    %\centering
    %\begin{tabular}{ccc}
    %    \hline
    %    Method & \acc   & \mnll\\
    %    \hline
    %    \textbf{\isgd} & 99.35 $\pm$ 0.05 & \textbf{0.0226} $\pm$ 0.0010 \\
    %    \textbf{\mcd} & \textbf{99.38} $\pm$ 0.02 & 0.0242 $\pm$ 0.0017 \\
    %    \textbf{\swag} & 99.14 $\pm$ 0.07 & 0.0291 $\pm$ 0.0012 \\
        % {} & {} & {}\\
    %    \hline
    %\end{tabular}
    % }
%\end{table}

%\begin{table}[!t]
   % \caption{Performance comparison of \textsc{ResNet-56} on \textsc{cifar10}.}
%    \label{tab:cifar10_results}
 %   \centering
        % \resizebox{\textwidth}{!}{
  %      \begin{tabular}{ccc}
   %     \hline
    %    Method & \acc   & \mnll\\
    %    \hline
    %    \textbf{\isgd} & 94.37 $\pm$ 0.15 & 0.2011 $\pm$ 0.0035 \\
        %\textbf{\isgd} & 94.07 $\pm$ 0.02 & 0.1949 $\pm$ 0.0046 \\
     %   \textbf{\mcd} & \textbf{95.15} $\pm$ 0.10 & 0.1734 $\pm$ 0.0033 \\
      %  \textbf{\swag} & 94.90 $\pm$ 0.08 & \textbf{0.1551} $\pm$ 0.0016 \\
        %\textbf{\baseline} & 94.32 $\pm$ 0.21 & 0.2045 $\pm$ 0.0032 \\
       % \hline
    %\end{tabular}
    % }
%\end{table}

We now use \textsc{ResNet-18}, \textsc{ResNet-56} and \textsc{VGG-16} networks on \textsc{cifar10}. For this set of experiments we compare \isgd, \mcd, \swag and \mandt, reporting test accuracy and \mnll. We also report a comparison between \isgd and \cyc using the results in  \cite{Zhang2020Cyclical} for \textsc{ResNet-18}, albeit the \mnll was not available.  These results are summarized in Tab.~\ref{tab:cifar10_results}, Tab.~\ref{tab:cifar10_vgg_results} and Tab.~\ref{tab:cifar10_res18_results}. At test time we use 30 samples for all methods. Results are obtained averaging over five independent seeds. \isgd is competitive with other methods. Moreover, \isgd is the the easiest to tune, as its learning rate is fixed and computed automatically.

\section{Conclusion}
\label{sec:conclusion}
\sg methods allowed Bayesian posterior sampling algorithms, such as \mcmc, to regain relevance in an age when data-sets have reached extremely large sizes. However, despite mathematical elegance and promising results, standard approaches from the literature are restricted to simple models. 
The sampling properties of these algorithms are determined by simplifying assumptions on the loss landscape, which do not hold for deep networks. \sgmcmc algorithms require vanishing learning rates, which force practitioners to develop creative annealing schedules that are often model specific and difficult to justify.

We have attempted to target these weaknesses by suggesting a simpler algorithm that relies on fewer parameters and mild assumptions compared to the literature. We introduced a unified mathematical notation to deepen our understanding of the role of the \sg noise and learning rate on the behavior of \sgmcmc algorithms. We presented a practical variant of the \sgd algorithm, which uses a constant learning rate, and an additional noise to perform Bayesian posterior sampling. Our proposal is derived from an ideal method, which guarantees that samples are generated from the true posterior. When the noise terms are empirically estimated, our method automatically determines the learning rate, and it offers a very good approximation to the posterior, as demonstrated by our experimental campaign.

% \pietro{Future work?}

% We verified empirically the quality of our approach, and compared its performance to state-of-the-art \sgmcmc and alternative methods. Results, which span a variety of settings, indicated that our method is competitive -- an in some cases superior -- to the alternatives from the state-of-the-art, while being much simpler to use, as it didn't require any learning rate tuning.

% We are currently investigating ways to make the estimator of $\lambda^{(p)}$ in our proposal more robust, especially in the case of deeper architectures. 
% On the theoretical side, we will investigate principled ways of relaxing the assumptions to achieve speedups in sampling efficiency.

\bibliographystyle{abbrvnat}
\bibliography{biblio}

\newpage
\clearpage
\newpage

\nocite*{}
\clearpage
\appendix
\section{\mcmc Through the Lenses of Langevin Dynamics}
\label{appa}
\subsection{The minibatch gradient approximation}
Starting from the gradient of the logarithm of the posterior density:
\begin{equation*}
   -\nabla f(\thetavect)=\sum\limits_{i=1}^N \nabla \log p(\boldsymbol{U}_i | \thetavect)+\nabla \log p(\thetavect),
\end{equation*}
it is possible to define its \textit{minibatch} version by computing the gradient on a random subset $\mathcal{I}_{N_b}$ with cardinality $N_b$ of all the indexes. The minibatch gradient $\boldsymbol{g}(\thetavect)$ is computed as 
\begin{equation*}
   -\boldsymbol{g}(\thetavect)=\frac{N}{N_b} \sum\limits_{i=1}^{N_b} \nabla \log p(\boldsymbol{U}_i | \thetavect)+\nabla \log p(\thetavect),
\end{equation*}
By simple calculations it is possible to show that the estimation is unbiased ($E(\boldsymbol{g}(\thetavect))=\nabla f(\thetavect)$). The estimation error covariance is defined to be $E\left[\left( \boldsymbol{g}(\thetavect)-\nabla f(\thetavect)\right)\left( \boldsymbol{g}(\thetavect)-\nabla f(\thetavect)\right)^{\top}\right]=2\Bmath$. 

If the minibatch size is large enough, invoking the central limit theorem, we can state that the minibatch gradient is normally distributed:
\begin{equation*}\label{gaussmb}
\boldsymbol{g}(\thetavect)\sim N( \nabla f(\thetavect), 2 \Bmath).
\end{equation*}

\subsection{Gradient methods without momentum}\label{sec:1stord}
\noindent \textbf{The \sde from discrete time} We start from the generalized updated rule of \sgd:
\begin{flalign*}
\boldsymbol{\delta} \thetavect_{n}=-\eta \boldsymbol{P}(\thetavect_{n-1})( \boldsymbol{g}(\thetavect_{n-1})+\boldsymbol{w}_{n}).
\end{flalign*}
Since $\boldsymbol{g}(\thetavect_{n-1})\sim N(\nabla f(\thetavect_{n-1}),2\boldsymbol{B}(\thetavect_{n-1}))$ we can rewrite the above equation as:
\begin{flalign*}
\boldsymbol{\delta} \thetavect_{n}=-\eta \boldsymbol{P}(\thetavect_{n-1})( \nabla f(\thetavect_{n-1})+\boldsymbol{w}^{'}_{n}),
\end{flalign*}
where $\boldsymbol{w}^{'}_{n}\sim N(0,2\boldsymbol{\Sigma}(\thetavect_{n-1}))$. If we separate deterministic and random component we can equivalently write:
\begin{flalign*}
\boldsymbol{\delta} \thetavect_{n}=-\eta \boldsymbol{P}(\thetavect_{n-1})\nabla f(\thetavect_{n-1})+\eta \boldsymbol{P}(\thetavect_{n-1})\boldsymbol{w}^{'}_{n}=
-\eta \boldsymbol{P}(\thetavect_{n-1})\nabla f(\thetavect_{n-1})+\sqrt{2\eta \boldsymbol{P}^2(\thetavect_{n-1})\boldsymbol{\Sigma}(\thetavect_{n-1})}\boldsymbol{v}_{n}
\end{flalign*}
where $\boldsymbol{v}_{n}\sim N(0,\eta\eye)$.
When $\eta$ is small enough ( $\eta\rightarrow dt$) we can interpret the above equation as the discrete time simulation of the following \sde \cite{gardiner2004handbook}: 
\begin{equation*}
d\thetavect_t=-\boldsymbol{P}(\thetavect_t)\nabla f(\thetavect_t) dt+\sqrt{2\eta \boldsymbol{P}(\thetavect_t)^2\boldsymbol{\Sigma}(\thetavect_t)}d\boldsymbol{W}_t,
\end{equation*}
where $d\boldsymbol{W}_t$ is a $d-$dimensional Brownian motion.

\noindent \textbf{Proof of \Cref{theo1}.} The stationary distribution of the above \sde, $\rho(\thetavect)\propto \exp(-\phi(\thetavect))$, satisfies the following \fpe:
\begin{flalign*}
0=\mathrm{Tr}\left\{
    \nabla \left[
        \nabla^\top \left(f(\thetavect)\right)\Pmath \rho(\thetavect)+\eta \nabla^\top(\Pmath^2\Sigmamath \rho(\thetavect))
        \right]
    \right\},
\end{flalign*}
that we rewrite as
\begin{flalign*}
0=\mathrm{Tr}\{
    \nabla[
        \nabla^\top \left(f(\thetavect)\right)\Pmath \rho(\thetavect)-\eta  \nabla^\top (\phi(\thetavect)) \Pmath^2\Sigmamath\rho(\thetavect)+\eta \nabla^\top(\Pmath^2\Sigmamath) \rho(\thetavect)
        ]
 \}.
\end{flalign*}
The above equation is verified with $\nabla f(\thetavect)=\nabla\phi(\thetavect)$ if
\begin{equation*}
\begin{cases}
\nabla^\top(\Pmath^2\Sigmamath)=\zerovect\\
\eta \Pmath^2\Sigmamath=\Pmath \rightarrow \eta \Pmath=\Sigmamath^{-1}
\end{cases}
\end{equation*}
that proves Theorem \ref{theo1}.

\subsection{Gradient methods with momentum}\label{sec:2ndord}
\noindent \textbf{The \sde from discrete time.} The general set of update equations for (discrete-time) momentum-based algorithms is:
\begin{flalign*}
\begin{cases}
\boldsymbol{\delta} \thetavect_{n}=\eta \boldsymbol{P}(\thetavect_{n-1})\boldsymbol{M}^{-1} \boldsymbol{r}_{n-1}\\
\boldsymbol{\delta} \boldsymbol{r}_{n}=-\eta \boldsymbol{A}(\thetavect_{n-1}) \boldsymbol{M}^{-1}\boldsymbol{r}_{n-1}-\eta \boldsymbol{P}(\thetavect_{n-1})( \boldsymbol{g}(\thetavect_{n-1})+\boldsymbol{w}_{n}).
\end{cases}
\end{flalign*}
Similarly to the case without momentum, we rewrite the second equation of the system as 
\begin{flalign*}
&\boldsymbol{\delta} \boldsymbol{r}_{n}=-\eta \boldsymbol{A}(\thetavect_{n-1}) \boldsymbol{M}^{-1}\boldsymbol{r}_{n-1}-\eta \boldsymbol{P}(\thetavect_{n-1})( \boldsymbol{g}(\thetavect_{n-1})+\boldsymbol{w}_{n})=-\eta \boldsymbol{A}(\thetavect_{n-1}) \boldsymbol{M}^{-1}\boldsymbol{r}_{n-1}-\eta \boldsymbol{P}(\thetavect_{n-1})\nabla f(\thetavect_{n-1})+\\&\sqrt{2\eta \boldsymbol{P}^2(\thetavect_{n-1})\boldsymbol{\Sigma}(\thetavect_{n-1})}\boldsymbol{v}_{n}
\end{flalign*}
where again $\boldsymbol{v}_{n}\sim N(0,\eta\eye)$.
If we define the super-variable  $\zetavect=\left[\thetavect,\boldsymbol{r}\right]^\top$, we can rewrite the system as: 
\begin{flalign*}
\boldsymbol{\delta}\zetavect_{n}= -\eta\begin{bmatrix} \zerovect & -\boldsymbol{P}(\thetavect_{n-1})\\ \boldsymbol{P}(\thetavect_{n-1}) & \boldsymbol{A}(\thetavect_{n-1}) \end{bmatrix}\boldsymbol{s}(\zetavect_{n-1})+\sqrt{2\eta \boldsymbol{D}(\zetavect_{n-1})}\boldsymbol{\nu}_{n}
\end{flalign*}
where $\boldsymbol{s}(\zetavect)=\begin{bmatrix}\nabla f(\thetavect)\\\boldsymbol{M}^{-1}\boldsymbol{r}\end{bmatrix}$, $\boldsymbol{D}(\zetavect)=\begin{bmatrix} \zerovect & \zerovect\\ \zerovect & \Pmath^2\Sigmamath\end{bmatrix}$ and $\boldsymbol{\nu}_{n}\sim N(0,\sqrt{\eta}\eye)$.

As the learning rate goes to zero ($\eta\rightarrow dt$), similarly to the previous case, we can interpret the above difference equation as a discretization of the following \fpe
\begin{flalign*}
 d\zetavect_t=-\begin{bmatrix} \zerovect & -\boldsymbol{P}(\thetavect_{t})\\ \boldsymbol{P}(\thetavect_{t}) & \boldsymbol{A}(\thetavect_{t}) \end{bmatrix}\boldsymbol{s}(\zetavect_{t})+\sqrt{2\eta \boldsymbol{D}(\zetavect_{t})}d\boldsymbol{W}_t
\end{flalign*}

\noindent \textbf{Proof of \Cref{theo2}.} As before we assume that the stationary distribution has form $\rho(\zetavect)\propto \exp(-\phi(\zetavect))$. The corresponding \fpe is:
\begin{flalign*}
0=\mathrm{Tr}\left(\nabla \left(\boldsymbol{s}(\zetavect)^{\top} \begin{bmatrix} \zerovect & -\boldsymbol{P}(\thetavect)\\\boldsymbol{P}(\thetavect) & \boldsymbol{A}(\thetavect)\end{bmatrix}\rho(\zetavect)+\eta\left(\nabla^{\top}\left(\boldsymbol{D}(\zetavect)\rho(\zetavect)\right)\right)\right)\right).
\end{flalign*}
Notice that since $\nabla^{\top} \boldsymbol{D}\left(z\right)=0$ we can rewrite: 
\begin{flalign*}
 &0=\mathrm{Tr}\left(\nabla \left(\boldsymbol{s}(\zetavect)^{\top} \begin{bmatrix} \zerovect & -\boldsymbol{P}(\thetavect)\\\boldsymbol{P}(\thetavect) & \boldsymbol{A}(\thetavect)\end{bmatrix}\rho(\zetavect)+\eta \nabla^{\top} (\rho(\zetavect)) \boldsymbol{D}(\zetavect) \right)\right)\\
 &=\mathrm{Tr}\left(\nabla \left(\boldsymbol{s}(\zetavect)^{\top} \begin{bmatrix} \zerovect & -\boldsymbol{P}(\thetavect)\\\boldsymbol{P}(\thetavect) & \boldsymbol{A}(\thetavect)\end{bmatrix}\rho(\zetavect)-\eta\nabla^{\top}(\phi(\zetavect)) \boldsymbol{D}(\zetavect)\rho(\zetavect)\right)\right)\\
 &=\mathrm{Tr}\left(\nabla \left(\boldsymbol{s}(\zetavect)^{\top} \begin{bmatrix} \zerovect & -\boldsymbol{P}(\thetavect)\\\boldsymbol{P}(\thetavect) & \boldsymbol{A}(\thetavect)\end{bmatrix}\rho(\zetavect)-\eta \nabla^{\top}(\phi(\zetavect))\begin{bmatrix} \zerovect & \zerovect\\ \zerovect & \boldsymbol{P}(\thetavect)^2\Sigmamath \end{bmatrix}\rho(\zetavect)\right)\right)
\end{flalign*}
that is verified with $\nabla \phi(\zetavect)=\boldsymbol{s}(\zetavect)$ if: 
\begin{equation*}\label{condmom2}
 \begin{cases}
\nabla^{\top} \boldsymbol{P}(\thetavect)=\zerovect\\
\Amath=\eta \boldsymbol{P}(\thetavect)^2\Sigmamath.
\end{cases}   
\end{equation*}
If $\nabla^{\top} \boldsymbol{P}(\thetavect)=\zerovect$, in fact: 
\begin{flalign*}
   & \mathrm{Tr}\left(\nabla \left(\nabla^{\top}(\phi(\zetavect))\rho(\zetavect)\begin{bmatrix} \zerovect & -\boldsymbol{P}(\thetavect)\\\boldsymbol{P}(\thetavect) & \zerovect \end{bmatrix}\right)\right)=\nabla^{\top}\left(\begin{bmatrix} \zerovect & -\boldsymbol{P}(\thetavect)\\\boldsymbol{P}(\thetavect) & \zerovect \end{bmatrix}\nabla(\phi(\zetavect))\rho(\zetavect)\right)=\\
  &  \nabla^{\top}\left(\begin{bmatrix} \zerovect & -\boldsymbol{P}(\thetavect)\\\boldsymbol{P}(\thetavect) & \zerovect \end{bmatrix}\right)\nabla(\phi(\zetavect))\rho(\zetavect)+\mathrm{Tr}\left(\begin{bmatrix} \zerovect & -\boldsymbol{P}(\thetavect)\\\boldsymbol{P}(\thetavect) & \zerovect \end{bmatrix}\nabla\left(\nabla^{\top}(\phi(\zetavect))\rho(\zetavect)\right)\right)=0,
\end{flalign*}

since $\nabla^{\top}\begin{bmatrix} \zerovect & -\boldsymbol{P}(\thetavect)\\\boldsymbol{P}(\thetavect) & \zerovect \end{bmatrix}=\zerovect$ and the second term is zero due to the fact that $\begin{bmatrix} \zerovect & -\boldsymbol{P}(\thetavect)\\\boldsymbol{P}(\thetavect) & \zerovect \end{bmatrix}$ is anti-symmetric while $\nabla\left(\nabla^{\top}(\phi(\zetavect))\rho(\zetavect)\right)$ is symmetric.

Thus we can rewrite: 
\begin{flalign*}
 &\mathrm{Tr}\left(\nabla \left(\boldsymbol{s}(\zetavect)^{\top} \begin{bmatrix} \zerovect & -\boldsymbol{P}(\thetavect)\\\boldsymbol{P}(\thetavect) & \boldsymbol{A}(\thetavect)\end{bmatrix}\rho(\zetavect)-\eta \nabla^{\top}(\phi(\zetavect))\begin{bmatrix} \zerovect & \zerovect\\ \zerovect & \boldsymbol{P}(\thetavect)^2\Sigmamath \end{bmatrix}\rho(\zetavect)\right)\right)=\\
  &\mathrm{Tr}\left(\nabla \left(\boldsymbol{s}(\zetavect)^{\top} \begin{bmatrix} \zerovect & -\boldsymbol{P}(\thetavect)\\\boldsymbol{P}(\thetavect) & \boldsymbol{A}(\thetavect)\end{bmatrix}\rho(\zetavect)- \nabla^{\top}(\phi(\zetavect))\begin{bmatrix} \zerovect & \zerovect\\ \zerovect & \eta\boldsymbol{P}(\thetavect)^2\Sigmamath \end{bmatrix}\rho(\zetavect)\right)\right)=\\
    &\mathrm{Tr}\left(\nabla \left(\boldsymbol{s}(\zetavect)^{\top} \begin{bmatrix} \zerovect & -\boldsymbol{P}(\thetavect)\\\boldsymbol{P}(\thetavect) & \boldsymbol{A}(\thetavect)\end{bmatrix}\rho(\zetavect)- \nabla^{\top}(\phi(\zetavect))\begin{bmatrix} \zerovect & \zerovect\\ \zerovect & \Amath \end{bmatrix}\rho(\zetavect)\right)\right)=\\
 &\mathrm{Tr}\left(\nabla \left(\left(\boldsymbol{s}(\zetavect)^{\top}-\nabla^{\top}(\phi(\zetavect))\right) \begin{bmatrix} \zerovect & -\boldsymbol{P}(\thetavect)\\\boldsymbol{P}(\thetavect) & \boldsymbol{A}(\thetavect)\end{bmatrix}\rho(\zetavect)\right)\right)=0
\end{flalign*}
 then, $\nabla \phi(\zetavect)=\boldsymbol{s}(\zetavect)$, proving Theorem \ref{theo2}.

\section{\isgd method proofs and details}
\label{appb}
\subsection{Proof of \Cref{theo_isgd} }\label{cor1proof}
The requirement $\Cmath\succeq 0 \quad \forall \thetavect$, ensures that the injected noise covariance is valid.  The composite noise matrix is equal to $\Sigmamath=\Lambdamath$. Since $\nabla^\top\Sigmamath=\nabla^\top\Lambdamath=\zerovect$ and $\eta \Pmath= \Lambdamath^{-1}$ by construction, then Theorem \ref{theo1} is satisfied.

\subsection{Proof of optimality of $\Lambdamath$}\label{lamopt}
Our design choice is to select $\lambda^{(p)} = \beta^{(p)}$. By the assumptions the matrix $\Bmath$ is diagonal, and consequently $\Cmath=\Lambdamath-\Bmath$ is diagonal as well. The preconditioner $\Lambdamath$ must be chosen to satisfy the positive semi-definite constraint, i.e. $\Cmath_{ii}\geq 0\quad \forall i,\forall \thetavect$. Equivalently, we must satisfy $\lambda^{(p)}-\mathbf{b}_j(\thetavect)\geq 0 \quad \forall j \in I_p,\forall p,\forall \thetavect$, where $I_p$ is the set of indexes of parameters belonging to $p_{th}$ layer. By assumption 3, i.e. $\beta^{(p)} = \max_{k\in I_p}\mathbf{b}_k(\thetavect)$, to satisfy the positive semi-definite requirement in all cases the minimum valid set of $\lambda^{(p)}$ is determined as $\lambda^{(p)} = \beta^{(p)}$.
%By assumption 3, i.e. $\beta^{(p)} = \sum_{k\in I_p}b_k(\thetavect)$, it is easy to show that $b_j(\thetavect), j\in I_p$, is upper bounded as $b_j(\thetavect)\leq \beta^{(p)}$. To satisfy the positive semi-definite requirement in all cases the minimum valid set of $\lambda^{(p)}$ is then determined as $\lambda^{(p)} = \beta^{(p)}$.

\subsection{Estimation of $\lambda^{(p)}$ }\label{lamest}

\noindent \textbf{The case of Gaussian \sg noise.} We here give additional details on the estimation of $\lambda^{(p)}$. The simple and naive estimation described in the paper is the following: $\lambda^{(p)}=\max_{j\in I_p}(\boldsymbol{g}_j(\thetavect)^{(p)})^2$. For the Gaussian \sg noise case we found however the following (safe) looser estimation of the maximum noise covariance to be more stable:
$\lambda^{(p)}=\sum\limits_{j\in I_p}b_j(\thetavect)=\frac{||\boldsymbol{g}(\thetavect)^{(p)}||^2}{2}$.  
From a practical point of view, we found the following filtering procedure to be useful and robust:
\begin{equation}
    \lambda^{(p)}\leftarrow \mu\lambda^{(p)}+(1-\mu)\frac{||\boldsymbol{g}^{(p)}(\thetavect)||^2}{2}
\end{equation}
where an exponential moving average is performed with estimation momentum determined by $\mu$. Notice that during sampling, the same smoothing can be applied to the tracking of $\Bmath$. We refer to the variant of \isgd implemented using this estimator as \textbf{\isgdG}. In this supplement we also considered the case of having a unique, and not layerwise, learning rate, that we indicate by justapposing the \textbf{(SLR)} acronym to the right of the methods. In this case, the unique equivalent $\lambda$ is computed as $\sum\limits_p \lambda^{(p)}$.

\noindent \textbf{The case of Heavy Tailed Noise.} 
A shared assumption of \sgmcmc methods is the Gaussianity of \sg noise. While this can be justified with the C.L.T.  for relatively simple models (linear models or simple feed-forward networks), this assumption has been challenged in the deep learning domain \cite{csimcsekli2019heavy,pmlr-v97-simsekli19a} suggesting that from complex architectures the noise distribution is heavy tailed. In particular, the hypothesis is that the noise follows and \textit{$\alpha$-stable} distribution, i.e.
\begin{equation}
    w\sim p(w)=\mathcal{F}^{-1}\left(\exp(-|ct|^{\alpha})\right)
\end{equation}
where $\alpha \in [0,2]$. Notice that except for particular cases, $p(w)$ can not be expressed in closed form. 
In general, when $\alpha<2$ the variance of the distribution goes to infinity and thus dealing with all methods that require the estimation or the usage of a covariance is tricky. It is interesting to underline that for $\alpha=2$ the distribution is the usual Gaussian one.

Having acknowledged that the noise is not Gaussian for deep models (at least) two possibilities can be considered: the first one is to study the \sde with L\`{e}vy Noise instead of Brownian, using a formalism similar to the one considered in \cite{csimvsekli2017fractional}, where \textit{fractional} \fpe have been considered. Several practical difficulties are however tied to this choice, such as the necessity to numerically approximate the fractional derivative of order $\alpha$ or the necessity to have full batch evaluations.
% (in \cite{csimvsekli2017fractional} mini-batches are also considered but the role of implicit noise is neglected). \giulio{I would like to avoid to give away for free this idea, I think this could be an interesting and unexplored future work.}

The second possibility, the one we used to present the results in the main paper and that we name \isgdAlpha in this supplement, is to neglect the fact that the noise is non-Gaussian, treat this as an approximation error, and use for the theoretical calculations the Gaussian distribution that is closest to the real noise distribution. In particular, for the one dimensional case, we minimize the $l_2$-distance between $p(x)$ and $q(x)$, where $p(x)=\sqrt{2\pi\sigma^2}\exp\left(-\frac{x^2}{2\sigma^2}\right)$ and $q(x)=\mathcal{F}^{-1}\left(\exp(-|ct|^{\alpha})\right)$. As stated above, in general no closed form exists for $q(x)$. Thanks to Parseval's equality, however, we can compute the distance in the frequency domain between the two distributions, i.e.
\begin{equation}
    C=\int\limits_{-\infty}^{+\infty} |p(x)-q(x)|^2 dx=\int\limits_{-\infty}^{+\infty} |p(t)-q(t)|^2 dt
\end{equation}
where $p(t)=\exp(-\frac{\sigma^2 t^2}{2})$ and $q(t)=\exp(-|ct|^{\alpha})$. Since we are optimizing w.r.t. $\sigma$, we can write the equivalent cost function
\begin{flalign}
    &C_{eq}=\int\limits_{-\infty}^{+\infty} |p(t)|^2 dt-2\int\limits_{-\infty}^{+\infty} p(t)q(t) dt=\int\limits_{-\infty}^{+\infty} \exp(-\sigma^2 t^2)dt -2 \int\limits_{-\infty}^{+\infty} \exp(-\frac{\sigma^2 t^2}{2})\exp(-|ct|^{\alpha})dt\nonumber\\&\frac{\sqrt{\pi}}{\sigma}-\frac{2}{\sigma}\int\limits_{-\infty}^{+\infty} \exp(-\frac{\tau^2}{2})\exp(-|\frac{c}{\sigma}\tau|^{\alpha})d\tau=\frac{\sqrt{\pi}}{\sigma}-\frac{2\sqrt{2\pi}}{\sigma}\int\limits_{-\infty}^{+\infty}\frac{1}{\sqrt{2\pi}} \exp(-\frac{\tau^2}{2})\exp(-|\frac{c}{\sigma}\tau|^{\alpha})d\tau\nonumber=\\&\frac{1}{\sigma}\left(\sqrt{\pi}-\sqrt{2\pi}E_{T\sim N(0,1)}[\exp\left(-|\frac{c}{\sigma}T|^{\alpha}\right)]\right).
\end{flalign} 
Equivalently, we can maximize for $r=\frac{c}{\sigma}$, the following function $r\left(\sqrt{\pi}-\sqrt{2\pi}E_{T\sim N(0,1)}[\exp\left(-|rT|^{\alpha}\right)]\right)$. The expected value does not have a closed form solution, but since the integral is single dimensional, it is possible to integrate numerically and derive the optimal $r$ for a given tail index, i.e. $\hat{r}=\arg\min C(r,\alpha)$ and consequently the optimal $\sigma$ as $\hat{\sigma}=\frac{c}{\hat{r}}$. Notice that even for moderately small values of $\alpha$ (i.e. $\alpha>0.5$), the optimal value is roughly $\frac{1}{\sqrt{2}}$, implying that a matching of the scales is sufficient: $\tilde{\sigma}^2=2c^2$. %In our implementation a simple lookup-table computed once is sufficient to derive the optimal $\sigma$ given knowledge of $\alpha,c$
%\begin{figure}
%    \centering
%    \includegraphics[scale=0.4]{figures/rvsalpha.jpg}
%    \caption{Caption}
%    \label{fig:rvsalpha}
%\end{figure}
%\giulio{Rephrase sentence}
%From a practical point of view, it is then sufficient to set $\tilde{\sigma}^2=2c^2$.
The parameters $\alpha,c$ are estimated (extending the results  of \cite{csimcsekli2019heavy,vehel2018explicit} to space varying settings) as described below.
Given a sequence of $N=N_1\times N_2$ samples $w[n]$ from an alpha-stable distribution, it is possible to estimate $\alpha,c$ using
\begin{flalign}
    &\frac{1}{\hat{\alpha}}=\frac{1}{\log(N_1)}\left(\frac{1}{N_2}\sum\limits_{i=0}^{N_2-1}\log\left|\sum\limits_{j=0}^{N_1-1} w[iN_1+j]\right| -\frac{1}{N}\sum\limits_{i=0}^{N-1}\log\left|w[i]\right|\right)\\
    &\hat{c}=\exp(\frac{1}{N}\sum\limits_{i=0}^{N-1}\log\left|w[i]\right|-\left(\frac{1}{\hat{\alpha}}-1\right)\gamma)
\end{flalign}
where $\gamma= 0.5772156649015329\dots$ is the Euler-Macheroni constant.
Notice that the computational cost for estimation of the two quantities is dominated by the calculation of logarithms, in fact for a full sequence of $N$ independent samples the cost is for the estimation of $\alpha$ $N+N_2$ absolute values, $N+N_2$ logarithms, $2N$ sums, with a per sample cost roughly equal to the cost of 1 logarithm evaluation, and for the estimation of $c$ the cost is $N$ logarithms, sums and absolute values (and thus similarly the cost is dominated by the log evaluation). When considering vectors of independent $d-$dimensional samples, the computational cost scales as $\mathcal{O}(d)$ logarithms.

Notice that for the \isgdalpha version we treated biases and weights of the layers as unique groups of parameters.

\noindent \textbf{Additional details on estimation.} 
Having chosen one of the two variants \isgdalpha or \isgdG for the estimation of $\lambda^{(p)}$, that we generically indicate as \isgdx, we still need to clarify what are the possibilites for the estimation of the parameters $\lambda^{p}$ before the sampling.
We considered three schemes:
\begin{itemize}
    \item \textbf{\isgda}: the starting point is a freshly initialized model. The estimation is performed while training, similarly to \cite{mandt2017stochastic} and \cite{ahn2012bayesian}, and a filtered version of the instantaneous estimation is stored;
    \item \textbf{\isgdb}: we start from a pre-trained model, and a warm-up phase is necessary. We \textit{continue} the training during the warm-up phase and collect a filtered version of estimates, as for the previous case.
    \item \textbf{\isgdc}: we start from a pre-trained model, and therefore a warm-up phase is necessary. We \textit{freeze} the network and estimate $\lambda^{(p)}$ using an adequeate number of mini-batches.
\end{itemize}
Summarizing, all the possible combinations are
\isgdALL. While not always the best performing, we found the \isgdalpha version the more stable across a large range of hyperparameters, and in the spirit of practicality, in the main paper we report only results obtained with this version.
%$KL(p(x)||q(x))$,
%where $p(x)=\sqrt{2\pi\sigma^2}\exp\left(-\frac{x^2}{2\sigma^2}\right)$ and $q(x)=\mathcal{F}^{-1}\left(\exp(-|ct|^{\alpha})\right)$,
%w.r.t. $\sigma$. We can rewrite the divergence as follows $KL(p||q)=-H(p)-\int p(x)\log q(x) dx=$ cost function is the following

%\section{Alpha Stable parameter estimation}
%\label{appc}
%\input{alphaestim.tex}

\section{Toy Model}
\label{appd}
Next, we consider a simple numerical example whereby it is possible to analytically compute the true posterior distribution.
We define a simple 1-D regression problem, in which we have $D$ trigonometric basis functions: $f(x) = \wvect^\top \cos(\omega x-\pi/4)$, where $\wvect \in \mathbb{R}^{D\times1}$ contains the weights of $D$ features and $\omega \in \mathbb{R}^{D\times1}$ is a vector of fixed frequencies.
We consider a Gaussian likelihood with variance $0.1$ and prior $p(\wvect) = \mathcal{N}(0, I_D)$; the true posterior over $\wvect$ is known to be Gaussian and it can be calculated analytically.

\begin{figure}
    \centering
    \pgfplotsset{width=8.5cm}
    \pgfplotsset{height=5.5cm}
    \input{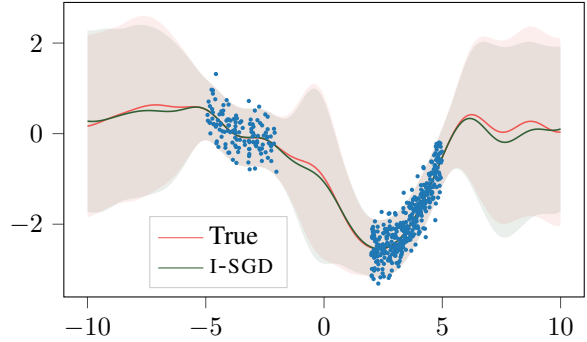}
    \caption{True and \isgd predictive posterior distributions on a simple example.}
    \label{fig:comparison}
\end{figure}

To assess the quality of the samples from the posterior obtained by \isgd, in Fig.~\ref{fig:comparison} we show the predictive posterior distribution (estimated using Eq.~\eqref{eq:predictive:montecarlo}) of \isgd, in comparison to the ``ground truth'' posterior. Visual inspection indicates that there is a good agreement between predictive posterior distributions, especially in terms of uncertainty quantification for test points far from the input training distribution.

\section{Experimental Methodology}
\label{appc}
We hereafter present additional implementation details and experiments. We variants of \isgd described in Section \ref{appb}. We report results for two preconditioner design choices: either tailored learning rates for each layer, as described in the main paper, or a unique learning rate where the equivalent $\lambda$ is the sum of all $\lambda^{(p)}$. We refer to this last version as single learning rate \slr.
\subsection{Regression tasks, with simple models}
At test time we use 100 samples to estimate the predictive posterior distribution, using Eq.~\eqref{eq:predictive:montecarlo}, for \isgd and \sghmc, and 10\,000 samples for \mcd. All our experiments use 10-splits. The considered batch size is 128 for all methods. In this set of experiments we use \isgdalphaa with $\mu=0.5$ during both warm-up and sampling. For both \isgdalphaa and \sghmc the warm-up has been set to 2000, and we do store a sample every 2000 iterations (the keepevery value is 2000).
\subsection{Classification task, simple \textsc{ConvNet}}
For the \textsc{LeNet-5} on \mnist  experiment, we consider \isgd variants, \mcd, and \swag. We moreover consider samples obtained by \sgd trajectories where the learning rate is derived as in \cite{mandt2017stochastic} (\mandt). At test time we use 30 samples for all methods. The batch size is 128, the temperature is $10^{-5}$ and the keepevery and warm-up periods are 100 and 100 respectively. For \swag we continued the training of the network for 5 epochs with learning rate 0.01.

We report results in Table \ref{tab:mnist_results_app}, in general all methods perform similarly.
\begin{table}[h!]
\centering
    \begin{tabular}{c c c}
        \hline
        Method & \acc   & \mnll\\
        \hline
        \textbf{\isgdAlphac} & \textbf{99.42} $\pm$ 0.03 & \textbf{0.0222} $\pm$ 0.0010 \\
        \textbf{\isgdGb} & 99.35 $\pm$ 0.05 & 0.0226 $\pm$ 0.0010 \\
        \textbf{\isgdGb \slr} & \textbf{99.42} $\pm$ 0.03 & \textbf{0.0222} $\pm$ 0.0014 \\
        \textbf{\mcd} & 99.38 $\pm$ 0.02 & 0.0242 $\pm$ 0.0017 \\
        \textbf{\swag} & 99.14 $\pm$ 0.07 & 0.0291 $\pm$ 0.0012 \\
        \textbf{\mandt} & 99.41 $\pm$ 0.03 & 0.0224 $\pm$ 0.0012 \\
        \hline
    \end{tabular}
    \caption{Performance comparison of \textsc{LeNet-5} on \mnist}
    \label{tab:mnist_results_app}
\end{table}

\subsection{Classification task, deeper models}
In the main paper we report results of \textsc{ResNet-56} on \cifar, using \isgdalphac,\swag, \mcd. Here we add results for \isgdGb,\isgdGc and \mandt. At test time we use 30 samples for all methods. For \isgdalphac the batch size is 64, temperature is $10^{-5}$, warm-up and keepevery are 800 and 4000 respectively. For the Gaussian noise implementations the batch size is 64, the estimation momentum $\mu$ is 0.9 and the keepevery and warm-up periods are 100 and 1000 respectively. For \swag we used the default parameters described in \cite{maddox2019simple}. Notice that for the \isgdGc version we treated biases and weights of the layers as unique groups of parameters.
We report results in Table \ref{tab:cifar10_results_app}. We notice that the various \isgd versions perform competitively and the \slr versions are worse in terms of performance.

We hereafter report additional results for \cifar   classification using \textsc{VGG-16}. We used the same parameters as for \textsc{ResNet-56}. We do omit results for \isgdGb because we encountered numerical problems. Results are reported in Table \ref{tab:cifar10_vgg_results}.

For \textsc{ResNet-18}, results in Table \ref{tab:cifar10_res18_results}, we use the same configuration as for the previous experiments: batch size is 64, temperature is $10^{-5}$, warm-up and keepevery are 800 and 4000 respectively.
\begin{table}[h!]
\centering
    \begin{tabular}{c c c}
        \hline
        Method & \acc   & \mnll\\
        \hline
        \textbf{\isgdAlphac} & 94.37 $\pm$ 0.15 & 0.2011 $\pm$ 0.0035\\
        \textbf{\isgdGb} & 94.07 $\pm$ 0.02 & 0.1949 $\pm$ 0.0046 \\
        \textbf{\isgdGb \slr} & 93.80 $\pm$ 0.21 & 0.2627 $\pm$ 0.0099 \\
        \textbf{\isgdGc} & 94.06 $\pm$ 0.08 & 0.1897 $\pm$ 0.0022 \\
        \textbf{\isgdGc \slr} & 94.39 $\pm$ 0.19 & 0.2027 $\pm$ 0.0037 \\
        \textbf{\mcd} & \textbf{95.15} $\pm$ 0.10 & 0.1734 $\pm$ 0.0033 \\
        \textbf{\swag} & 94.90 $\pm$ 0.08 & \textbf{0.1551} $\pm$ 0.0016 \\
        \textbf{\mandt} & 93.82 $\pm$ 0.19 & 0.2664 $\pm$ 0.0100 \\
%        \textbf{\baseline} & 94.32 $\pm$ 0.21 & 0.2045 $\pm$ 0.0032 \\
        \hline
    \end{tabular}
    \caption{Performance comparison of \textsc{ResNet-56} on \textsc{cifar10}}
    \label{tab:cifar10_results_app}
\end{table}

\begin{table}[h!]
\centering
    \begin{tabular}{c c c}
        \hline
        Method & \acc   & \mnll\\
        \hline
        \textbf{\isgdAlphac} & 92.73 $\pm$ 0.07 & 0.3577 $\pm$ 0.0046 \\
        \textbf{\isgdGb \slr} & 92.93 $\pm$ 0.10 & 0.3136 $\pm$ 0.0085 \\
        \textbf{\isgdGc} & 92.94 $\pm$ 0.11 & 0.2644 $\pm$ 0.0068 \\
        \textbf{\isgdGc \slr} & 92.91 $\pm$ 0.08 & 0.3389 $\pm$ 0.0063 \\
        \textbf{\mcd} & 92.71 $\pm$ 0.12 & 0.2470 $\pm$ 0.0067 \\
        \textbf{\swag} & \textbf{93.66} $\pm$ 0.13 & \textbf{0.1946} $\pm$ 0.0036 \\
        \textbf{\mandt} & 93.02 $\pm$ 0.06 & 0.3313 $\pm$ 0.0062 \\
  %      \textbf{\baseline} & 93.07 $\pm$ 0.10 & 0.3302 $\pm$ 0.0065 \\
        \hline
    \end{tabular}
    \caption{Performance comparison of \textsc{VGG-16} on \textsc{cifar10}}
    \label{tab:cifar10_vgg_results}
\end{table}

\begin{table}[h!]
\centering
    \begin{tabular}{c c c}
        \hline
        Method & \acc   & \mnll\\
        \hline
        \textbf{\isgdAlphac} & 95.38 $\pm$ 0.12 & 0.1794 $\pm$ 0.0044 \\
        \cyc & 95.73 $\pm$ 0.03 & N/A \\
  %      \textbf{\baseline} & 93.07 $\pm$ 0.10 & 0.3302 $\pm$ 0.0065 \\
        \hline
    \end{tabular}
    \caption{Performance comparison of \textsc{ResNet-18} on \textsc{cifar10}}
    \label{tab:cifar10_res18_results}
\end{table}

\end{document}